\documentclass[sigconf]{acmart}

\renewcommand\footnotetextcopyrightpermission[1]{}  
\usepackage[switch]{lineno} 
\usepackage{xcolor}         
\usepackage{microtype}
\usepackage{graphicx}
\usepackage{subfigure}
\usepackage{booktabs} 

\usepackage{hyperref}
\usepackage[ruled,vlined]{algorithm2e}
\usepackage{pifont}
\newcommand{\cmark}{\ding{51}} 
\newcommand{\xmark}{\ding{55}} 
\usepackage{amsmath}
\usepackage{amssymb}
\usepackage{mathtools}
\usepackage{amsthm}
\usepackage{array}     
\usepackage{multirow}  

\theoremstyle{plain}
\newtheorem{theorem}{Theorem}[section]

\newtheorem{lemma}[theorem]{Lemma}

\theoremstyle{definition}

\theoremstyle{remark}

\AtBeginDocument{%
  }

\begin{document}


\title{An Uncertainty-aware DETR Enhancement Framework for Object Detection}

\author{Xingshu Chen$^{*}$}
\email{chenxsh53@mail2.sysu.edu.cn}
\affiliation{%
  \institution{Sun Yat-sen University}
  \city{}
  \state{}
      \country{}
}

\author{Sicheng Yu$^{*}$}
\email{yusch@mail2.sysu.edu.cn}
\affiliation{
 \institution{AI Thrust, HKUST(GZ)}
 \city{}
 \state{}
 \country{}
}

\author{Chong Cheng}
\email{ccheng735@connect.hkust-gz.edu.cn}
\affiliation{
 \institution{AI Thrust, HKUST(GZ)}
 \city{}
 \state{}
 \country{}
}

\author{Hao Wang$^{\dagger}$}
\email{haowang@hkust-gz.edu.cn}
\affiliation{%
 \institution{AI Thrust, HKUST(GZ)}
 \city{}
 \state{}
 \country{}}

\author{Ting Tian$^{\dagger}$}
\email{tiant55@mail.sysu.edu.cn}
\affiliation{%
 \institution{Sun Yat-sen University}
 \city{}
 \state{}
 \country{}}
 
\thanks{$^{*}$ Equal contribution.}
\thanks{$^{\dagger}$ Corresponding authors.}





\begin{abstract}

This paper investigates the problem of object detection with a focus on improving both the localization accuracy of bounding boxes and explicitly modeling prediction uncertainty. Conventional detectors rely on deterministic bounding box regression, ignoring uncertainty in predictions and limiting model robustness.
In this paper, we propose an uncertainty-aware enhancement framework for DETR-based object detectors. We model bounding boxes as multivariate Gaussian distributions and incorporate the Gromov-Wasserstein distance into the loss function to better align the predicted and ground-truth distributions. Building on this, we derive a Bayes Risk formulation to filter high-risk information and improve detection reliability. We also propose a simple algorithm to quantify localization uncertainty via confidence intervals.
Experiments on the COCO benchmark show that our method can be effectively integrated into existing DETR variants, enhancing their performance. We further extend our framework to leukocyte detection tasks, achieving state-of-the-art results on the LISC and WBCDD datasets. These results confirm the scalability of our framework across both general and domain-specific detection tasks. Code page: \url{https://github.com/ParadiseforAndaChen/An-Uncertainty-aware-DETR-Enhancement-Framework-for-Object-Detection}.
\end{abstract}








\maketitle

\pagestyle{plain}
\fancyhead{}  

\section{Introduction}
\label{intro}
Object detection aims to tackle the problems of bounding box regression and object classification for each object of interest. Classical convolution-based detectors \cite{redmon2016yolo1, tian2020fcos, wang2023yolov7, wang2024yolov10, wang2024gold, wang2025yolov9}, along with recently proposed Transformer-based end-to-end detectors \cite{carion2020DETR, zhu2020Deformable-DETR,dai2021dynamic-DETR, zhang2022dino-DETR, cai2023align-DETR, jia2023H-DETR, pu2024rank-DETR,hou2025relation-DETR}, have significantly advanced the performance of object detection.

\begin{figure}[t]
\vskip 0.2in
\begin{center}
\centerline{\includegraphics[width=\columnwidth]{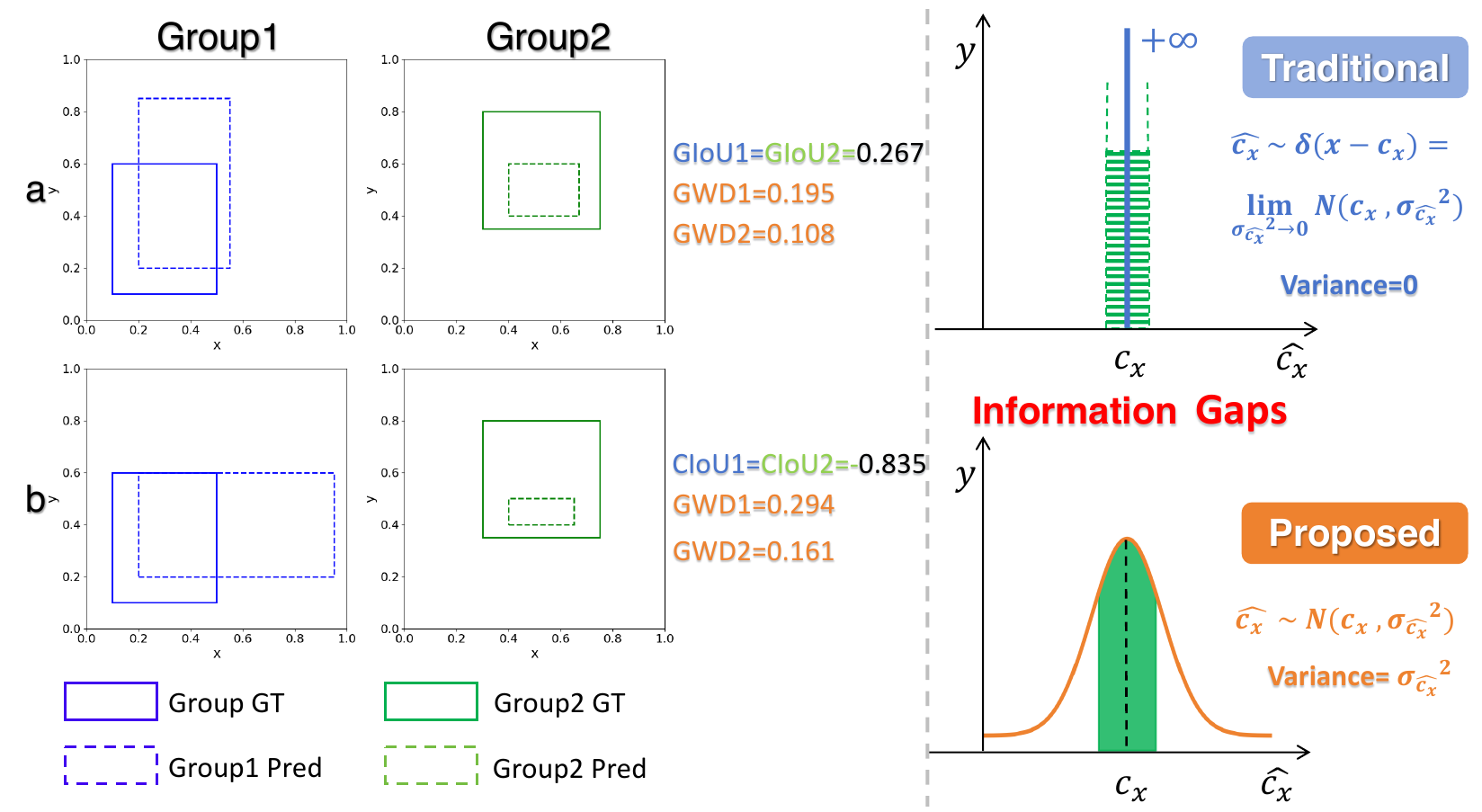}}
\caption{The left side shows two examples (Ex. a, Ex. b) where GIoU and CIoU fail to distinguish completely different ground truth and prediction pairs, while GW distance does.  On the right, taking center coordinate $c_x$ as an example, traditional methods model bounding boxes as fixed values following a Dirac delta distribution, whereas we model them as Gaussian distributions with variance.}
\label{fig:Drawback_IOU}
\end{center}
\vskip -0.3in
\end{figure}

Despite these advancements, several challenges limit the performance and reliability of object detection models. One major challenge lies in the \textbf{formulation of bounding box regression}. In traditional object detection frameworks, bounding boxes are represented by fixed coordinates and dimensions \cite{ren2016faster,liu2016ssd,tian2020fcos}. Model training typically relies on L1 loss and IoU-based losses (e.g., GIoU \cite{rezatofighi2019GIoU}, DIoU and CIoU \cite{zheng2020DCIoU}), which measure geometric similarity based on these fixed representations. However, these geometrically approximated objectives cause discontinuous gradients and hinder stable convergence, while failing to provide a systematic quantification of uncertainty. Figure~\ref{fig:Drawback_IOU} provides two examples where GIoU and CIoU produce identical values for pairs of completely different ground truth and predicted bounding boxes. This suggests that deterministic 2D geometric representations fail to capture sufficient information in predictions.

From a probabilistic perspective, representing bounding boxes with fixed coordinates is equivalent to modeling predictions as Dirac delta distributions, the limiting case of Gaussian distributions when variance approaches zero. This formulation does not account for prediction uncertainty. While some prior works \cite{choi2019gaussian, qiu2020offset, wang2021NWD, yang2021GWD-old} have modeled bounding boxes as Gaussian distributions, their fixed variance still fails to capture the uncertainty in predictions. A new approach to model bounding boxes is essential to overcome these limitations and address the performance bottleneck in object detection models.

Another challenge lies in \textbf{quantifying localization uncertainty}. The confidence score reflects the model's certainty about object classification. However, there is a lack of a reasonable characterization of overall uncertainty in localization. Some works \cite{he2019bounding, li2021generalized, lee2022localization} have provided uncertainty in the four directions of the predicted bounding box, but they do not account for the overall uncertainty in localization. Unlike classification scores, object detection still lacks a measure of the overall reliability of localization.

Given its end-to-end design and growing influence in modern object detection, DETR has emerged as a strong baseline. To address the above issues, we propose a novel uncertainty-aware DETR enhancement framework. Specifically, we establish a one-to-one correspondence between the ground truth and 2D Gaussian distributions, and model the predictions as a 4D Gaussian distribution with a learnable covariance matrix. This formulation effectively captures prediction uncertainty, providing more information for model training. To measure the discrepancy between distributions across different dimensions, we introduce the Gromov-Wasserstein (GW) distance \cite{delon2022gromovgwd-ours}. By minimizing the GW distance, we ensure that the predicted and ground truth distributions become statistically closer, thereby improving prediction accuracy and model's robustness. Furthermore, we provide a theoretical upper bound that characterizes the convergence of the Gromov-Wasserstein distance to zero.

Leveraging the statistical properties of these distributions, we derive the formulation of Bayes Risk for bounding box regression, representing the theoretical lower bound of the regression loss achievable by the model. This Bayes Risk is then incorporated into DETRs to refine the internal modules. By filtering high-risk predictions, the model focuses more on reliable outputs, thereby improving performance. Finally, we propose a distribution-based algorithm to characterize overall localization uncertainty. The algorithm constructs prediction confidence intervals to provide a solid measure of uncertainty for the predicted boxes.

To evaluate the generalizability of our framework, we conduct experiments in both general and domain-specific settings. On the COCO benchmark, our method can be seamlessly integrated into various DETR variants, resulting in improved detection performance. To assess its applicability to specialized tasks, we extend our framework to leukocyte detection—a classic medical imaging task. In this context, providing reliable estimates of prediction uncertainty is particularly valuable, as it can assist clinicians in making more informed diagnostic decisions, thereby carrying significant clinical importance. 
Experiments on the WBCDD and LISC datasets demonstrate that our method outperforms state-of-the-art cell detection models while offering interpretable uncertainty estimates. \textbf{The contributions of this paper are summarized as follows:}

\begin{enumerate}
    \item We propose modeling bounding boxes as multivariate Gaussian distributions with learnable covariance matrices to capture uncertainty, and introduce the Gromov-Wasserstein distance for distribution alignment. 
    
    \item We derive Bayesian risk minimization for DETR-based detectors and introduce a confidence interval algorithm that quantifies localization uncertainty, enabling risk-aware detection.

    \item Our method integrates seamlessly into existing DETR variants, improving detection performance on COCO benchmark and achieving state-of-the-art results on LISC and WBCDD datasets for leukocyte detection, demonstrating strong generalization across both general and specific domains.

\end{enumerate}

\section{Related Work}

\textbf{Bounding Box Modeling and Metric. }Traditional methods treat bounding boxes as fixed coordinates, using IoU-based metrics to capture the geometric similarity between predictions and ground truth. IoU is the most widely used metric; however, it is only effective when bounding boxes have overlap. GIoU \cite{rezatofighi2019GIoU} addresses non-overlapping cases by introducing a penalty term based on the smallest enclosing box. However, when one bounding box completely contains another, GIoU degenerates to IoU. To overcome this issue, CIoU and DIoU \cite{zheng2020DCIoU} incorporate additional factors such as the overlapping area, central point distance, and aspect ratio, covering more scenarios. Building on this, SIoU \cite{gevorgyan2022SIoU} further accounts for the angle between bounding boxes. Despite these extensions, such modeling ignores prediction uncertainty, and IoU-based metrics still face significant limitations.

Recent works have modeled bounding boxes as probabilistic distributions and introduced distribution-based metrics. \citet{wang2021NWD} and \citet{yang2021GWD-old} represent bounding boxes as 2D Gaussian distributions. The former introduced the Normalized Wasserstein Distance to alleviate the sensitivity of IoU to location deviations in tiny objects, while the latter proposed Gaussian Wasserstein Distance to address boundary discontinuity and the square-like problem in oriented object detection. However, these works fail to capture prediction uncertainty. In our approach, we model the ground truth as a 2D Gaussian distribution and the prediction as 4D Gaussian distribution, where the variance measures prediction uncertainty. To compare distributions of different dimensions, we introduce the Gromov-Wasserstein distance \cite{delon2022gromovgwd-ours} as a metric.

\textbf{Localization Uncertainty. }Localization uncertainty in object detection refers to the model's ability to estimate the confidence or uncertainty associated with predicted bounding box locations. \citet{lakshminarayanan2017simple} and \citet{harakeh2020bayesod} use Monte Carlo dropout within a Bayesian framework to account for prediction uncertainty, improving model performance. \citet{he2019bboxuncertainty} estimate bounding box uncertainty by minimizing the KL-divergence between the Gaussian distribution of the predicted bbox and the Dirac delta distribution of the ground truth bbox on Faster R-CNN \cite{ren2016faster}. \citet{lee2022localization} propose Uncertainty-Aware Detection (UAD), equipping FCOS \cite{tian2020fcos} with a localization uncertainty estimator that reflects box quality along four directions of the predicted bbox. However, Monte Carlo dropout is computationally expensive, and these methods only estimate uncertainty in four directions, failing to capture the overall localization uncertainty of the entire box. Additionally, they are tailored to CNN-based architectures and cannot be directly applied to Transformer-based models like DETR.

\textbf{DETR for Object Detection. }The pioneering work DETR \cite{carion2020DETR} introduced an end-to-end transformer-based framework \cite{vaswani2017transformer} for 2D object detection, inspiring numerous follow-up studies. For example, Deformable DETR \cite{zhu2020Deformable-DETR} tackled scalability issues by adopting deformable attention, enabling efficient processing of high-resolution images without sacrificing accuracy. Conditional DETR \cite{meng2021conditional-DETR} refined query initialization to improve detection accuracy. DINO-DETR \cite{zhang2022dino-DETR} introduced a query denoising scheme to accelerate convergence. H-DETR \cite{jia2023H-DETR} proposed a hybrid matching strategy that combines one-to-one matching with auxiliary one-to-many matching to enhance training efficiency. Additionally, Relation-DETR \cite{hou2025relation-DETR} incorporated positional relation priors as attention biases, improving both interpretability and detection performance, and achieving SOTA results on multiple benchmarks.  These methods have collectively advanced DETR’s performance across diverse object detection tasks. Our approach, in contrast, provides a general and flexible framework that integrates seamlessly with these methods.

\section{Distribution Modeling and Theoretical Analysis}

In this section, we detail the process of modeling bounding boxes as distributions and discuss the advantages. To measure the discrepancy between prediction and ground truth, we introduce the GW distance and provide a theoretical proof of its convergence property. Additionally, we derive the Bayes Risk based on those distributions to further refine modules in DETR-based models.

\subsection{Distribution Modeling of Bounding Boxes }
In bounding box (bbox) regression tasks, the training objective is to make the predicted bbox as similar as possible to the ground truth bbox. This makes the formulation of bboxes and the measurement of their ``similarity'' critical to the success of the model. Traditionally, bboxes are represented as fixed coordinates,  formulated as Dirac delta distributions. However, this approach only captures precise boundary information and ignores the inherent uncertainty in predictions, leading to limitations in training process.

To address this, we model bboxes as Gaussian distributions, enabling a probabilistic perspective to measure and align the ground truth and prediction. This generalizes Dirac delta distributions, which can be seen as the limiting case of Gaussian distributions as the variance approaches zero. Specifically, the ground truth bbox is modeled as a 2D Gaussian distribution, derived by back-projecting its inscribed ellipse. We treat the predicted bbox's as a 4D Gaussian distribution, given that the model's outputs consist of four components. As training progresses, the two distributions become increasingly similar, leading to more accurate  predictions.

\begin{figure}[t]
\begin{center}
\centerline{\includegraphics[width=0.8\columnwidth]{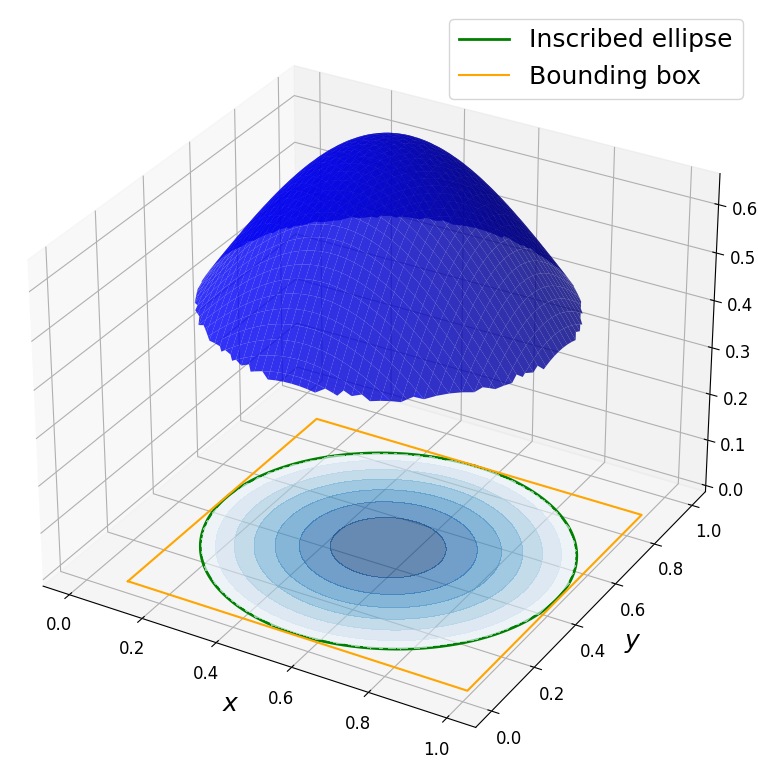}}
\vspace{-5pt}
\caption{Portion of the entire whole Gaussian surface that satisfies the condition $(\mathbf{x} - \boldsymbol{\mu}_g)^\top \boldsymbol{\Sigma}_g^{-1} (\mathbf{x} - \boldsymbol{\mu}_g) \leq 1$ and its projection onto the coordinate plane.}
\label{GT_projection}
\end{center}
\vspace{-20pt}
\end{figure}

\begin{figure*}[t]
\begin{center}
\centerline{\includegraphics[width=0.95\textwidth]{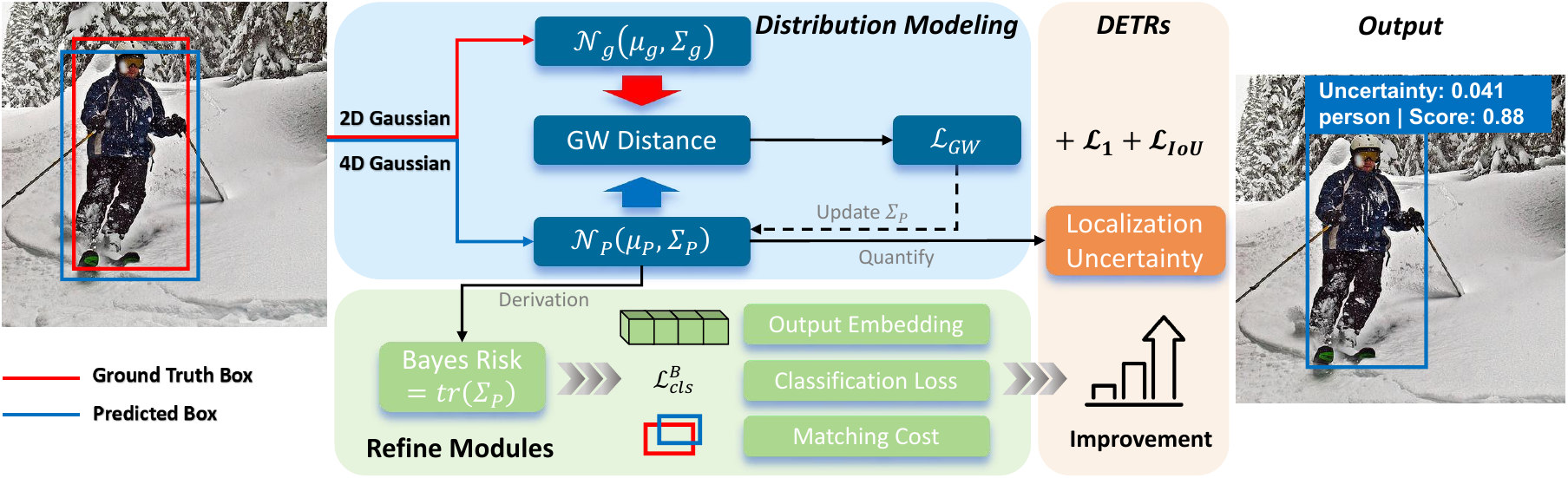}}
\caption{\textbf{The diagram illustrates our approach to modeling bounding boxes and integrating it into DETR-based frameworks.} Ground truth and predicted bounding boxes are modeled as 2D and 4D gaussian distributions respectively, with the Gromov-Wasserstein Distance serving as the loss. Based on the the distribution of the prediction, we derive the Bayes Risk of the predicted bbox and use it to refine three modules in existing DETRs frameworks. Additionally, using Gaussian-based modeling, we quantify the localization uncertainty of the prediction. The final output includes the target class, classification score, and localization uncertainty.}
\label{whole_pipline}
\end{center}
\vskip -0.3in
\end{figure*}

\textbf{2D Gaussian Distribution for Ground Truth. }
A ground truth bounding box \( R = (c_x, c_y, w, h) \), where \( c_x \) and \( c_y \) represent the center coordinates while \( w \) and \( h \) denote the width and height, contains both foreground and background pixels. Foreground pixels are primarily concentrated within the inscribed ellipse, while background pixels distribute across the remaining regions. Let \( \textbf{x} = (x, y)^T \), \( \boldsymbol{\mu}_{g} = (c_x, c_y)^T \), and \(\boldsymbol{\Sigma}_{g} = \text{Diag}\left(\frac{w^2}{4}, \frac{h^2}{4}\right)\). The equation of the inscribed ellipse can be expressed as:
\begin{align}
&(\textbf{x} - \boldsymbol{\mu}_{g})^\top \boldsymbol{\Sigma}_{g}^{-1} (\textbf{x} - \boldsymbol{\mu}_{g}) = 1.
\end{align}

We back-project the inscribed ellipse into 3D space, resulting in a surface of 2D Gaussian distribution $\mathcal{N}_{g}(\boldsymbol{\mu}_{g}, \boldsymbol{\Sigma}_{g})$, whose density function is given by:
\begin{align}
f(\textbf{x} | \boldsymbol{\mu}_g, \boldsymbol{\Sigma}_g) &= 
\frac{\exp\left(-\frac{1}{2} (\textbf{x} - \boldsymbol{\mu}_g)^\top \boldsymbol{\Sigma}_g^{-1} (\textbf{x} - \boldsymbol{\mu}_g)\right)}
{2\pi \lvert \boldsymbol{\Sigma}_g \rvert^{\frac{1}{2}}}.
\end{align}As shown in \autoref{GT_projection}, the blue surface represents the Gaussian surface region satisfying $(\mathbf{x} - \boldsymbol{\mu}_g)^\top \boldsymbol{\Sigma}_g^{-1} (\mathbf{x} - \boldsymbol{\mu}_g) \leq 1$. Its projection onto the coordinate plane corresponds to the inscribed ellipse of the ground truth bounding box. This establishes a one-to-one correspondence between the ground truth bounding box and the 2D Gaussian distribution.


\textbf{4D Gaussian Distribution for Prediction. }When predicting the bounding box of an object, the model outputs four values $\hat{R} = (\hat{c}_x, \hat{c}_y, \hat{w}, \hat{h})$, which represent fixed location information. However, predictions inherently involve uncertainty. We assume that each component of $\hat{R}$ follows a 1D Gaussian distribution $\mathcal{N}_{i}({\mu_{i}}, {\sigma}_i^2)$, where $i = \{\hat{c}_x, \hat{c}_y, \hat{w}, \hat{h}\}$, $\mu_i = \{c_x, c_y, w, h\}$, and $0 < \sigma_i^2 \leq 1$. Thus, the predicted bounding box $\hat{R} = (\hat{c}_x, \hat{c}_y, \hat{w}, \hat{h})$ can be modeled as a 4D Gaussian distribution $\mathcal{N_{P}}(\boldsymbol{\mu}_\mathcal{P}, \boldsymbol{\Sigma}_\mathcal{P})$, where 
\[
\boldsymbol{\mu}_\mathcal{P} =
\begin{bmatrix}
c_x \\
c_y \\
w \\
h
\end{bmatrix},
\quad
\boldsymbol{\Sigma}_\mathcal{P} =
\begin{bmatrix}
\sigma_{\hat{c}_x}^2 & \sigma_{\hat{c}_x \hat{c}_y}^2 & \sigma_{\hat{c}_x \hat{w}}^2 & \sigma_{\hat{c}_x \hat{h}}^2 \\
\sigma_{\hat{c}_y \hat{c}_x}^2 & \sigma_{\hat{c}_y}^2 & \sigma_{\hat{c}_y \hat{w}}^2 &  \sigma_{\hat{c}_y \hat{h}}^2 \\
\sigma_{\hat{w} \hat{c}_x}^2 & \sigma_{\hat{w} \hat{c}_y}^2 & \sigma_{\hat{w}}^2 & \sigma_{\hat{w} \hat{h}}^2 \\
\sigma_{\hat{h} \hat{c}_x}^2 & \sigma_{\hat{h} \hat{c}_y}^2 & \sigma_{\hat{h} \hat{w}}^2 & \sigma_{\hat{h}}^2
\end{bmatrix}.
\]
For simplicity, we assume that the components of 
$\hat{R}$ are independent, so that $\boldsymbol{\Sigma}_\mathcal{P} = \text{Diag}(\sigma_{\hat{c}_x}^2, \sigma_{\hat{c}_y}^2, \sigma_{\hat{w}}^2,\sigma_{\hat{h}}^2)$. The standard deviation $\sigma_i$ is a learnable parameter, measuring uncertainty of the estimation. As $\sigma_i$ approaches 0, the model becomes more confident in its predictions.

\subsection{GW Distance for Bounding Box Regression }
After modeling the bounding boxes as Gaussian distributions, a metric is required to measure the difference between them. The Gromov-Wasserstein (GW) distance \cite{delon2022gromovgwd-ours} provides a way to compare distributions in different dimensions. Given $m=\mathcal{N}_g(\boldsymbol{\mu}_g, \boldsymbol{\Sigma}_g)$ and $n=\mathcal{N_{P}}(\boldsymbol{\mu}_\mathcal{P}, \boldsymbol{\Sigma}_\mathcal{P})$, the GW distance in this case is defined as:
\begin{align}
&GW_2^2(m, n) = \\
&\inf_{\pi \in \Pi(m, n)} \iint   
\left( \|x - x'\|^2 - \|y - y'\|^2 \right)^2 
\mathrm{d}\pi_1 \ \mathrm{d}\pi_2. \notag
\end{align}where $\pi_1=\pi(x,y),\pi_2=\pi(x',y')$. The analytical solution for the GW distance in this case will be provided in \ref{Analytical solution}. To establish the convergence property of the GW distance, we present the following theorem:

\begin{theorem}
\label{thm:convergence}
Let  $
\boldsymbol{\Sigma}_*=
\begin{pmatrix}
\boldsymbol{\Sigma}_g& \boldsymbol{0} \\
\boldsymbol{0} & \boldsymbol{0}
\end{pmatrix} \in \mathbb{R}^4
$, \( \boldsymbol{\Sigma}_\mathcal{P} = \boldsymbol{\Sigma}_* +\Delta \boldsymbol{\Sigma} \), then as $\|\Delta \boldsymbol{\Sigma}\|_F \to 0$:     \[
    GW_2^2(m, n) = O( {\|\Delta \boldsymbol{\Sigma}\|}^2_F).
\]

\end{theorem}\autoref{thm:convergence} implies that as $ \|\Delta \boldsymbol{\Sigma}\|_F $ approaches zero, the square of the GW distance converges to zero at a rate proportional to\( \|\Delta \boldsymbol{\Sigma}\|_F^2 \). For a detailed proof, see \ref{proof of convergence}.

\subsection{Bayes Risk Derivation}
Bayes Risk represents the minimum achievable expected loss for the model. In our work, we compute the Bayes Risk for bounding box predictions to refine the modules in DETRs models, further enhancing performance. Given a loss function, the Bayes Risk is defined as:

\begin{align}
\textit{Risk}^* &= \inf_{\hat{R}} \mathbb{E}[\text{loss}(\hat{R}, R)].
\end{align} 

We derive the Bayes Risk based on the $L_2$ loss, as it provides a smoother optimization process and effectively captures large prediction errors. Using $L_2$ loss, the Bayes Risk is expressed as the following simplified form:
\begin{align}
\label{eq:risk_trace}
\textit{Risk}^* &= \sigma_{\hat{c}_x}^2 + \sigma_{\hat{c}_y}^2 + \sigma_{\hat{w}}^2 + \sigma_{\hat{h}}^2.
\end{align} which is just the trace of $\boldsymbol{\Sigma}_\mathcal{P}$. For a detailed derivation, please refer to \ref{Bayes Risk derivation}.

\section{Approach}

\autoref{whole_pipline} provides an overview of our approach to model the bounding boxes and enhance DETRs frameworks. First, we model the bounding boxes as Gaussian distributions and use the GW distance to measure their difference, serving as the loss for DETRs. Next, based on the distribution modeling, we derive the Bayes Risk and use it to refine modules within the DETRs. Finally, we calculate the localization uncertainty of the predicted bounding box.

\subsection{Loss Formulation Based on GW Distance}
For the bounding box regression problem, previous works in the DETRs primarily employed IoU-based Loss and $L_1$ Loss \cite{carion2020DETR}. However, as discussed in \autoref{intro}, these losses have certain limitations. We model the ground truth and predicted bounding boxes as Gaussian distributions and use the GW distance to measure their differences. Therefore, the bounding box regression loss in our work can be expressed as:
\begin{align}
    \label{box loss}
    \mathcal{L}_{box}= &\lambda_{iou}\mathcal{L}_{iou}(R,\hat{R})+\lambda_{L1}||R-\hat{R}||_1  \notag
    \\ &+\lambda_{gw}GW_2^2(\mathcal{N}_{g},\mathcal{N}_\mathcal{P}).
\end{align}
By introducing GW distance, the model gains a more comprehensive perspective by considering the distribution, which helps guide parameter optimization and enhances performance. Furthermore, as shown in \ref{Analytical solution}, the GW distance formulation includes the predicted covariance matrix, allowing its distribution parameters to be optimized for alignment. This, in turn, serves as the foundation for deriving Bayes Risk to further refine the modules.

\subsection{Bayes Risk Refinement Modules}
According to \autoref{eq:risk_trace}, the Bayes Risk of the the prediction is equal to the trace of $\boldsymbol{\Sigma}_\mathcal{P}$. For DETRs, we define the Normalized Bayes Risk vector $\mathcal{T}$ as:
\begin{align}
\label{eq:vector}
\mathcal{T} = \{t_1, t_2, \dots, t_N\} \in \mathbb{R}^{1 \times N}.
\end{align}
where $t_i = \textit{Risk}^*/4 $ represents normalized Bayes Risk for each object query in the decoder. $\mathcal{T}$ reflects the minimum expected loss made by the model. Using this information, we can refine different modules within DETRs, thereby improving its performance.

\textbf{Output Embedding. }
In DETRs, the embeddings $\mathcal{Z} \in \mathbb{R}^{d \times N}$ output by the transformer decoder are passed through a feedforward network to produce the final predictions. The term $1-\mathcal{T}$ reflects the confidence of predictions, where higher values indicate greater stability and lower error rates. We apply this term to refine these embeddings, followed by an additional MLP layer:
\begin{align}
\label{eq:refine_Z}
\mathcal{Z}_{\text{re}} &= \text{MLP}(\mathcal{Z} \odot (1- \mathcal{T})).
\end{align} where $\odot $ is the Hadamard product. By incorporating refined embeddings, the model gains confidence-aware representations, which enable more informed inference and reduce the impact of uncertain predictions. These refined embeddings help prioritize lower Bayes Risk predictions, allowing the model to focus on more reliable outputs.

\textbf{Classification Loss. }
Previous work \cite{cai2023align-DETR,pu2024rank-DETR} highlight that the misalignment between classification scores and localization accuracy limits the performance of DETRs. To address this, they incorporate IoU score $u$ and classification score $s$ into a unified term $r$ within the BCE loss, termed the IoU-aware Classification Loss:
\begin{align}
\label{eq:cls_loss}
\mathcal{L}_{cls} &= \sum_{i}^{N_{pos}} \text{BCE}(s_i, r_i) + \sum_{j}^{N_{neg}} s_j^2 \text{BCE}(s_j, 0).
\end{align} where $r=\left(\frac{\text{GIoU}(\hat{R}, R) + 1}{2} - s\right)^2$. A predicted bounding box with a high IoU score $u$ should correspond to a low Bayes Risk. Therefore, we extend the IoU-aware Classification Loss by weighting $r$ with Bayes Risk. Let $w = exp(-\textit{Risk}^*/4)$, the Bayes Risk aware BCE loss is defined as: 
\begin{align}
\label{eq:cls_loss}
\mathcal{L}_{cls}^{B} &= \sum_{i}^{N_{pos}} \text{BCE}(s_i, w_ir_i) + \sum_{j}^{N_{neg}} s_j^2 \text{BCE}(s_j, 0).
\end{align}By leveraging the Bayes Risk weighting mechanism, we effectively downweight the impact of lower-quality predictions while amplifying the influence of more accurate ones.

\textbf{Matching Cost. }
DETRs typically use the Hungarian algorithm for one-to-one matching. However, the matching cost, which simply sums the classification and regression costs, overlooks the relationship between the classification score $s$ and IoU $u$. Additionally, the linear representation of IoU cannot capture subtle variations when its value is high. To address these issues, inspired by \cite{pu2024rank-DETR}, we adopt a multiplicative form and a higher-order representation of IoU. Incorporating information from Bayes Risk, we propose a Bayes Risk refine matching cost:
\begin{align}
\mathcal{L}_{\text{match}}^{\text{Bayes Risk}} = s^{1+\textit{Risk}^*/4} \cdot u^{4+\textit{Risk}^*}.
\end{align}

\vspace{1pt}

\subsection{Quantify Localization Uncertainty.}
The predicted bounding box $\hat{R} = (\hat{c}_x, \hat{c}_y, \hat{w}, \hat{h})$ follows a 4D Gaussian distribution $\mathcal{N_{P}}(\boldsymbol{\mu}_\mathcal{P}, \boldsymbol{\Sigma}_\mathcal{P})$ , where $\boldsymbol{\mu}_\mathcal{P}=(c_x, c_y, w, h)^T$ and $\boldsymbol{\Sigma}_\mathcal{P} = \text{Diag}(\sigma_{\hat{c}_x}^2, \sigma_{\hat{c}_y}^2, \sigma_{\hat{w}}^2,\sigma_{\hat{h}}^2)$. Based on this distribution, the $95\%$ confidence intervals for $(\hat{c}_x, \hat{c}_y, \hat{w}, \hat{h})$ can be derived individually. Building on this, we propose an algorithm to compute the uncertainty of the predicted bounding box, as detailed in \autoref{alg:localization_uncertainty}. Section \ref{section:Localization Uncertainty Reliability} confirm that the proposed algorithm provides highly valuable uncertainty estimates, accurately reflecting the precision of the predicted bounding boxes.

\begin{algorithm}
   \caption{Quantify Localization Uncertainty}
   \label{alg:localization_uncertainty}
   \fontsize{9.5}{11}\selectfont  
   \setlength{\baselineskip}{10pt}  
   {\bfseries Input:} Predicted bounding box $\hat{R} = (\hat{c}_x, \hat{c}_y, \hat{w}, \hat{h})$, number of divisions $k$ \\
   {\bfseries Output:} Localization uncertainty

   1. Compute 95\% confidence intervals:
   \[
   [\hat{c}_x \pm 1.96\sigma_{\hat{c}_x}],  [\hat{c}_y \pm 1.96\sigma_{\hat{c}_y}],  [\hat{w} \pm 1.96\sigma_{\hat{w}}],  [\hat{h} \pm 1.96\sigma_{\hat{h}}]
   \]
   2. Divide each interval into $k$ equal parts:
   \[
   \{\hat{c}_{x_i}\}, \{\hat{c}_{y_i}\}, \{\hat{w}_i\}, \{\hat{h}_i\}, \quad i = 1, \dots, k
   \]
   3. Form $k$ bounding boxes:
   \[
   \hat{R}_i = (\hat{c}_{x_i}, \hat{c}_{y_i}, \hat{w}_i, \hat{h}_i), \quad i = 1, \dots, k
   \]
   4. Compute IoUs between $\hat{R}$ and each $\hat{R}_i$:
   \[
   \text{IoU}_i = \text{IoU}(\hat{R}, \hat{R}_i), \quad i = 1, \dots, k
   \]
   5. Calculate average IoU of top 5 values and get uncertainty:
   \vspace{-5pt} 
   \[
   \text{AvgIoU} = \frac{1}{5} \sum_{j=1}^5 \text{Top-5 IoU}_j,
   \text{Uncertainty} = 1 - \text{AvgIoU}
   \]

\vspace{-5pt}
\end{algorithm}
\vspace{-10pt}

\section{Experiment}

\definecolor{darkgreen}{rgb}{0.0, 0.5, 0.0} 

\begin{table*}[t] 
\caption{\textbf{Comparison of baseline models and those enhanced by our approach on COCO \texttt{val2017}.} All models were trained for 12 epochs. For Relation-DETR, the default classification loss was used without Bayes Risk modification due to constraints imposed by the model's structure. }
\label{main_results}
\vspace{-2pt}
\begin{center}
\fontsize{10}{11}\selectfont  
\setlength{\tabcolsep}{4.8pt} 
\renewcommand{\arraystretch}{1} 
\begin{tabular}{lccccccccc} %
\toprule
Model    & Backbone          & AP        & AP$_{50}$ & AP$_{75}$ & AP$_S$ & AP$_M$ & AP$_L$ \\
\midrule
H-DETR    & Res-50             & 48.7 & 66.4 & 52.9 & 31.2 & 51.5 & 63.5 \\

H-DETR$+$ours   & Res-50       & 50.1 \textcolor{darkgreen}{(\textbf{+1.4})} & 67.6 \textcolor{darkgreen}{(\textbf{+1.2})} & 54.8 \textcolor{darkgreen}{(\textbf{+1.9})} & 33.1 \textcolor{darkgreen}{(\textbf{+1.9})} & 53.8 \textcolor{darkgreen}{(\textbf{+2.3})} & 64.0 \textcolor{darkgreen}{(\textbf{+0.5})} \\

\midrule

H-DETR   & Swin-T     & 50.6 & 68.9 & 55.1 & 33.4 & 53.7 & 65.9 \\

H-DETR$+$ours   & Swin-T        & 51.7 \textcolor{darkgreen}{(\textbf{+1.1})} & 69.1 \textcolor{darkgreen}{(\textbf{+0.2})} & 56.6 \textcolor{darkgreen}{(\textbf{+1.5})} & 35.0 \textcolor{darkgreen}{(\textbf{+1.6})} & 55.0 \textcolor{darkgreen}{(\textbf{+1.3})} & 66.8 \textcolor{darkgreen}{(\textbf{+0.9})} \\

\midrule
DINO-DETR      & Res-50        & 49.0 & 66.6 & 53.5 & 32.0 & 52.3 & 63.0 \\
DINO-DETR$+$ours     & Res-50        & 50.2 \textcolor{darkgreen}{(\textbf{+1.2})} & 67.8 \textcolor{darkgreen}{(\textbf{+1.2})} & 54.9 \textcolor{darkgreen}{(\textbf{+1.4})} & 33.4 \textcolor{darkgreen}{(\textbf{+1.4})} & 53.6 \textcolor{darkgreen}{(\textbf{+1.3})} & 64.6 \textcolor{darkgreen}{(\textbf{+1.6})} \\
\midrule
Relation-DETR   & Res-50      & 51.7 & 69.1 & 56.3 & 36.1 & 55.6 & 66.1 \\
Relation-DETR$+$ours   & Res-50        & 51.9 \textcolor{darkgreen}{(\textbf{+0.2})} & 69.3 \textcolor{darkgreen}{(\textbf{+0.2})} & 56.6 \textcolor{darkgreen}{(\textbf{+0.3})} & 35.9 \textcolor{gray}{(\textbf{-0.2})} & 55.7 \textcolor{darkgreen}{(\textbf{+0.1})} & 66.7 \textcolor{darkgreen}{(\textbf{+0.6})} \\
\midrule
Relation-DETR   & Swin-L       & 57.8 & 76.1 & 62.9 & 41.2 & 62.1 & 74.4 \\
Relation-DETR$+$ours   & Swin-L      & 57.9 \textcolor{darkgreen}{(\textbf{+0.1})} & 76.2 \textcolor{darkgreen}{(\textbf{+0.1})} & 63.1 \textcolor{darkgreen}{(\textbf{+0.2})} & 41.8 \textcolor{darkgreen}{(\textbf{+0.6})} & 62.2 \textcolor{darkgreen}{(\textbf{+0.1})} & 74.2 \textcolor{gray}{(\textbf{-0.2})} \\

\bottomrule
\end{tabular}
\end{center}
\vspace{-2pt}
\end{table*}

\label{experiment}
\subsection{Experiment Setting}

To validate the effectiveness of our method in enhancing general DETR-based detectors and quantifying localization uncertainty, we conduct experiments on the COCO benchmark \cite{lin2014cocodata}. We select three representative DETR variants---H-DETR \cite{jia2023H-DETR}, DINO-DETR \cite{zhang2022dino-DETR}, and Relation-DETR \cite{hou2025relation-DETR}, and extend them with our approach. Each model is implemented with either ResNet-50 \cite{he2016resnet} or Swin Transformer \cite{liu2021swin-transformer} backbone. All models are trained on COCO \texttt{train} set with the standard 1x schedule and evaluated on the \texttt{val} set.

To further demonstrate the applicability of our method in biomedical scenarios, we evaluate it on two public datasets for leukocyte detection and classification: the Leukocyte Images for Segmentation and Classification (LISC) dataset \cite{rezatofighi2011LISC} and the White Blood Cell Detection Dataset (WBCDD) . Both datasets contain five types of white blood cells—neutrophils (NEU), eosinophils (EOS), monocytes (MON), basophils (BAS), and lymphocytes (LYM). We apply our enhanced H-DETR variant and compare it with several classic object detectors, including Faster R-CNN \cite{ren2016faster}, SSD \cite{liu2016ssd}, RetinaNet \cite{lin2017RetinaNet}, DETR \cite{carion2020DETR}, and Deformable DETR \cite{zhu2020Deformable-DETR}, as well as specialized leukocyte detection models such as TE-YOLOF \cite{xu2022TE-YOLOF}, YOLOv5-ALT \cite{guo2023YOLOv5-ALT}, and MFDS-DETR \cite{chen2024MFDS-DETR}. All models are trained on the respective training sets and evaluated on the test sets.

\subsection{Main Results}

\begin{figure}[t]
\begin{center}
\centerline{\includegraphics[width=0.97\columnwidth]{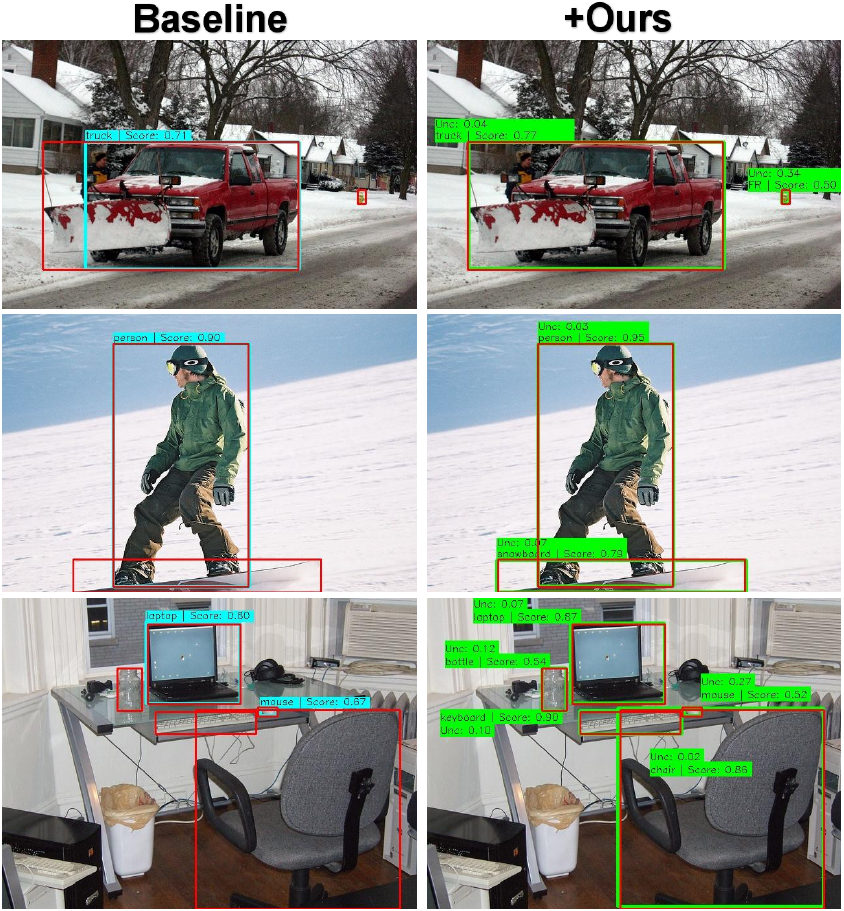}}
\caption{\textbf{Comparison of the results after integrating our method into H-DETR.} Red boxes denote ground truth, while blue and green boxes indicate predictions. Our method improves bounding box accuracy and enables DETRs to detect small, edge-blurred, or low-texture objects that the original model missed. Additionally, our method provides localization uncertainty, where more accurate predictions correspond to lower uncertainty.}
\label{dingxing}
\end{center}
\vspace{-15pt}
\end{figure}

\label{main result}
\textbf{Enhancing DETR-based Models. }\autoref{main_results} compares the performance of baseline models and those enhanced by our approach on the COCO \texttt{val2017} dataset. The results demonstrate that our approach consistently improves DETRs across different backbones. Specifically, for H-DETR, our method boosts AP to 50.1\%(\textbf{+1.4\%})with ResNet-50 and 51.7\%(\textbf{+1.1\%}) with Swin-Tiny backbone. Similarly, our enhancement improves DINO-DETR by \textbf{1.2\%} with ResNet-50. For Relation-DETR, which already sets new state-of-the-art performance in object detection, our approach further refines its results. With ResNet-50 and Swin-Large, AP increases by \textbf{0.2\%} and \textbf{0.1\%}, respectively. Although this improvements appear modest compared to previous results, they demonstrate the generalizability of our method in enhancing all already well-designed architectures.

It should be noted that our approach not only improves overall performance but also enables the model to detect challenging objects. \autoref{dingxing} provides illustrative examples, where H-DETR, after our enhancement, successfully detects previously missed objects, including a small fire hydrant, an edge-blurred snowboard, and a low-texture glass bottle.

\textbf{Application to Leukocyte Detection. } As shown in \autoref{table:application result}, our method outperforms previous approaches on both the LISC and WBCDD datasets. Compared with the current state-of-the-art MFDS-DETR, our framework improves overall detection accuracy by \textbf{+1.4}\% AP on LISC and \textbf{+1.9}\% AP on WBCDD. Our model also achieves higher precision on challenging categories such as Monocytes and Eosinophils, particularly in the WBCDD dataset. While some smaller classes may not reach the top scores, our method maintains consistently strong performance across all five leukocyte types in both datasets, demonstrating better generalization and robustness than other methods.

\begin{figure}[t]
\begin{center}
\centerline{\includegraphics[width=0.8\columnwidth]{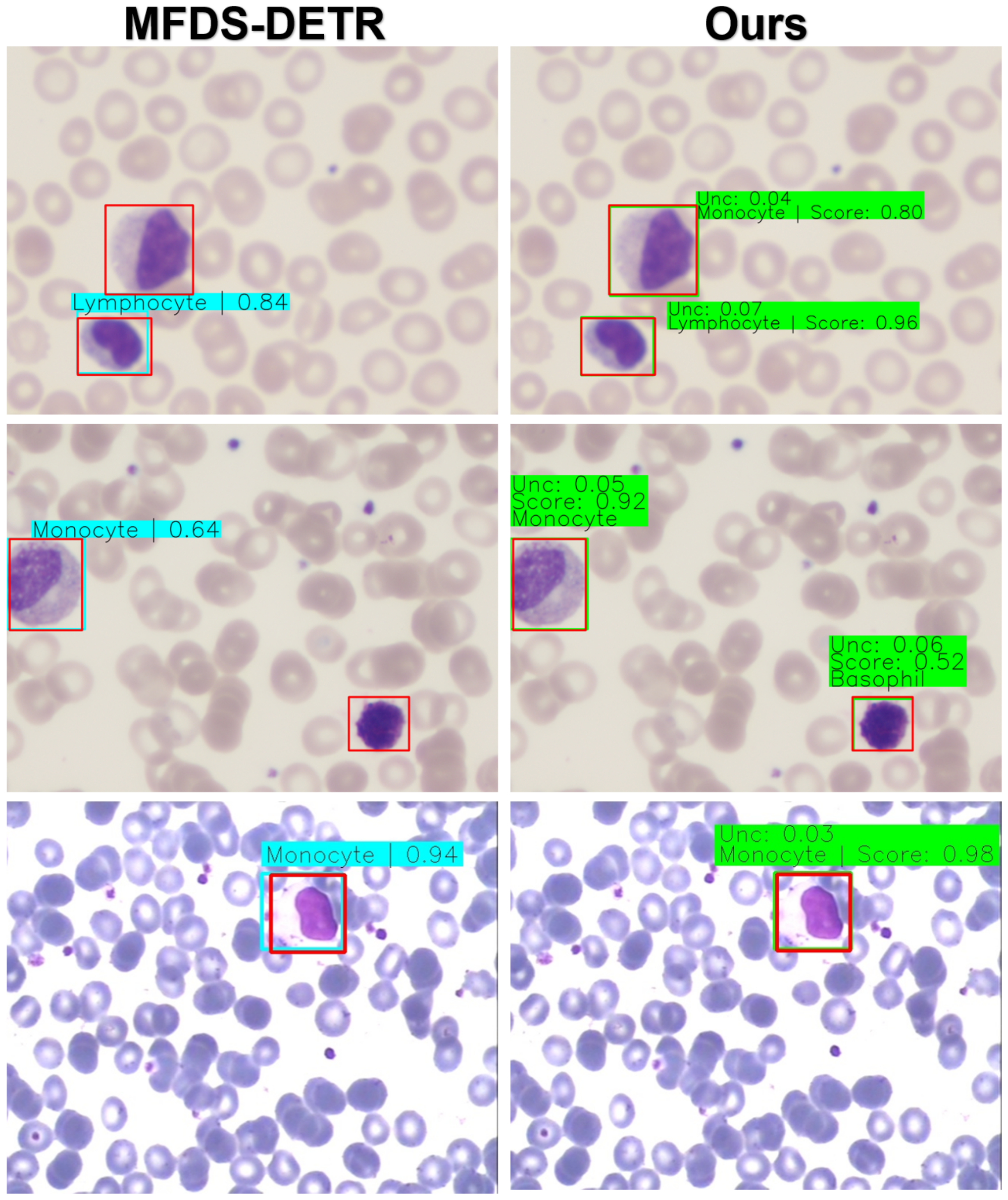}}
\vspace{-1pt}
\caption{Qualitative comparison with the state-of-the-art method. Our model detects more challenging leukocyte objects, produces more accurate bounding boxes, and provides meaningful localization uncertainty estimates.}
\label{dingxing2}
\end{center}
\vspace{-20pt}
\end{figure}

As illustrated in \autoref{dingxing2}, our model detects more ambiguous or overlapping cells and produces more accurate bounding boxes. These results confirm the effectiveness of our framework in extending to domain-specific biomedical object detection tasks.

\begin{table*}[t]
\centering
\caption{Comparison of leukocyte detection performance on the LISC and WBCDD datasets. Our method achieves state-of-the-art overall detection accuracy and maintains consistent performance across all five leukocyte subtypes.}
\vspace{-3pt}
\fontsize{8.5}{10}\selectfont  
\setlength{\tabcolsep}{0.7pt} 
\renewcommand{\arraystretch}{1.3} 
    \begin{tabular}{>{\centering\arraybackslash}p{3cm}|*{3}{>{\centering\arraybackslash}p{0.8cm}}*{5}{>{\centering\arraybackslash}p{0.85cm}}|*{3}{>{\centering\arraybackslash}p{0.8cm}}*{5}{>{\centering\arraybackslash}p{0.85cm}}}
    \toprule
    \multirow{2}{*}{\textbf{Method}} & \multicolumn{8}{c}{\textbf{LISC}}& \multicolumn{8}{c}{\textbf{WBCDD}}  \\ \cmidrule(lr){2-9} \cmidrule(lr){10-17}  
                        & AP & AP$_{50}$ & AP$_{75}$ & AP$_{\text{NEU}}$ & AP$_{\text{MON}}$ & AP$_{\text{EOS}}$ & AP$_{\text{LYM}}$ & AP$_{\text{BAS}}$
                        & AP & AP$_{50}$ & AP$_{75}$ & AP$_{\text{NEU}}$ & AP$_{\text{MON}}$ & AP$_{\text{EOS}}$ & AP$_{\text{LYM}}$ & AP$_{\text{BAS}}$ \\ \hline
    \textbf{Faster R-CNN} \cite{ren2016faster}& 76.5 & \textbf{100} & 96.9 & \underline{83.3} & 71.4 & 80.2 & 70.2 &77.5 & 58.2 & 73.7 & 72.4 & 84.9 & 53.1 & 41.5 & 73.1& 38.2\\ 
    \textbf{SSD} \cite{liu2016ssd}   & 70.3& 96.1& 92.7& 73.7& 72.0& 61.5& 68.9& 75.3& 64.2 &80.5 &77.9 &83.1 &48.0 &49.0 &67.2 &73.9\\ 
    \textbf{RetinaNet} \cite{lin2017RetinaNet}& 37.0 &52.1 &47.6 &46.1 &22.0 &19.7 &69.8 &19.7 &47.6 &57.0 &55.3 &85.1 &47.3 &31.1 &66.4 &7.9\\ 
    \textbf{DETR} \cite{carion2020DETR}& 77.8 &98.9 &\underline{98.9} &82.1 &76.0 &80.6 &72.5 &77.7 &66.8 &86.4 &82.5 &84.1 &53.4 &52.4 &73.6 &70.5\\ 
    \textbf{Deformable DETR} \cite{zhu2020Deformable-DETR}& 78.1 &\textbf{100.0} &95.4 &79.6 &77.2 &82.6 &72.9 &78.2 &74.9 &94.4 &93.6 &84.2 &68.7 &74.5 &73.7 &73.1\\ 
    \textbf{TE-YOLOF} \cite{xu2022TE-YOLOF} & 77.4 &\textbf{100.0} &94.9 &81.2 &79.2 &81.1 &69.3 &76.2 & 68.5 &88.7 &86.5 &86.9 &59.3 &69.3 &\underline{79.7} &47.2 \\ 
    \textbf{YOLOv5-ALT} \cite{guo2023YOLOv5-ALT} & 75.9 &98.8 &97.9 &\textbf{84.3} &74.5 &\underline{84.1} &\textbf{77.1} &59.5 &71.3 &\underline{98.2} &93.4 &\textbf{88.2} &62.7 &74.2 &72.2 &59.4 \\
    \textbf{MFDS-DETR} \cite{chen2024MFDS-DETR}& \underline{79.5} &\underline{99.9} &98.7 &75.2 &\underline{81.8} &83.9 &74.2 &\textbf{82.5} &\underline{79.7} &97.2 &\underline{96.8} &\underline{87.1} &\underline{71.5} &\underline{85.0} &\textbf{80.3} &\underline{74.9}\\
    \hline
    \textbf{Ours} & \textbf{80.9} & \textbf{100.0} & \textbf{100} &81.3 &\textbf{82.4} &\textbf{84.6} &\underline{75.3} &\underline{80.7} &\textbf{81.6} &\textbf{98.3} &\textbf{98.3} &85.6 &\textbf{76.6} &\textbf{89.0} &78.8 &\textbf{78.0}  \\
    \bottomrule
    \end{tabular}
    \label{table:application result}
\end{table*}

\subsection{Localization Uncertainty Reliability}
\label{section:Localization Uncertainty Reliability}
\begin{figure}[t]
\begin{center}
\centerline{\includegraphics[width=0.75\columnwidth]{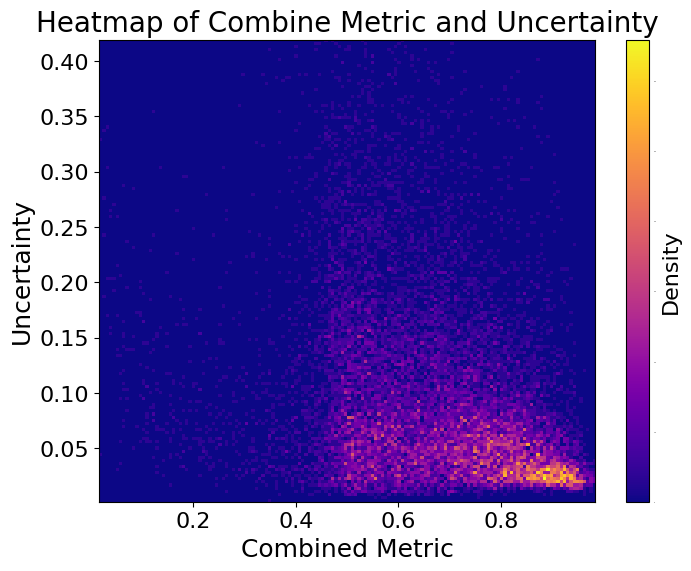}}
\vspace{-2pt}
\caption{\textbf{Heatmap between the Combined Metric and localization uncertainty.} $\text{Combined Metric} = s \cdot u^{0.5}$}
\label{uncertainty reliability}
\end{center}
\vspace{-2pt}
\end{figure}

During inference, our algorithm provides localization uncertainty to quantify the reliability of predicted bounding boxes. Ideally, an accurate prediction should have both a high classification score $s$ and a high IoU $u$ with the ground truth, corresponding to lower localization uncertainty.  Inspired by \cite{cai2023align-DETR,pu2024rank-DETR}, we use the following metric to jointly consider $s$ and $u$: $\text{Combined Metric} = s \cdot u^{0.5}$ as a measure of prediction quality. We then analyze its relationship with uncertainty. As shown in \autoref{uncertainty reliability}, lower Combined Metric values correspond to higher and more dispersed uncertainty, whereas higher values result in lower, more concentrated uncertainty, validating the reliability of our uncertainty estimation.

\subsection{Ablation Study}We conducted a series of experiments to evaluate the impact of each component in our approach on COCO benchmark. Using H-DETR with IoU-aware loss and a ResNet-50 backbone as the baseline, we progressively add and remove modules to demonstrate their effects. 
Note that GW distance is a prerequisite for introducing the Bayes Risk Refinement Modules.

\textbf{Gromov-Wasserstein Distance. }The GW distance measures the discrepancy between the ground truth distribution and predicted distribution. By minimizing the GW distance, we enhance prediction performance while optimizing the covariance matrix for Bayes Risk computation, which in turn refines subsequent modules for further improvements. As shown in \autoref{ablation_1}, incorporating GW distance as a loss term slightly improves the AP of the baseline model, with more notable gains in AP$_S$ and AP$_M$. However, since all subsequent modules rely on it as a prerequisite, its importance is further underscored. Appendix \ref{sec:appendix_gwd} presents a comparison of GW distance with other traditional metrics.

\begin{table}[t]
\caption{Ablation study on integrating GW distance as a loss function and the Bayes Risk Refinement Module (BRRM) into the baseline model.}
\vspace{-1pt}
\label{ablation_1}
\begin{center}
\fontsize{9.5}{11}\selectfont  
\setlength{\tabcolsep}{2pt} 
\renewcommand{\arraystretch}{1} 
\begin{tabular}{cccccccc} 
\toprule
GWD &  BRRM & AP & AP$_{50}$ & AP$_{75}$ & AP$_S$ & AP$_M$ & AP$_L$ \\
\midrule
\xmark  & \xmark & 49.2 & 66.7 & 53.9 & 31.7 & 52.4 & 63.7 \\
\cmark  & \xmark & 49.3 & 66.9 & 53.6 & 32.4 & 52.7 & 63.7 \\
\cmark  & \cmark & \textbf{50.1} & \textbf{67.6} & \textbf{54.8} & \textbf{33.1} & \textbf{53.8} & \textbf{64.0} \\
\bottomrule
\end{tabular}
\end{center}
\vspace{-3pt}
\end{table}

\begin{figure}[t]
\begin{center}
\centerline{\includegraphics[width=0.9\columnwidth]{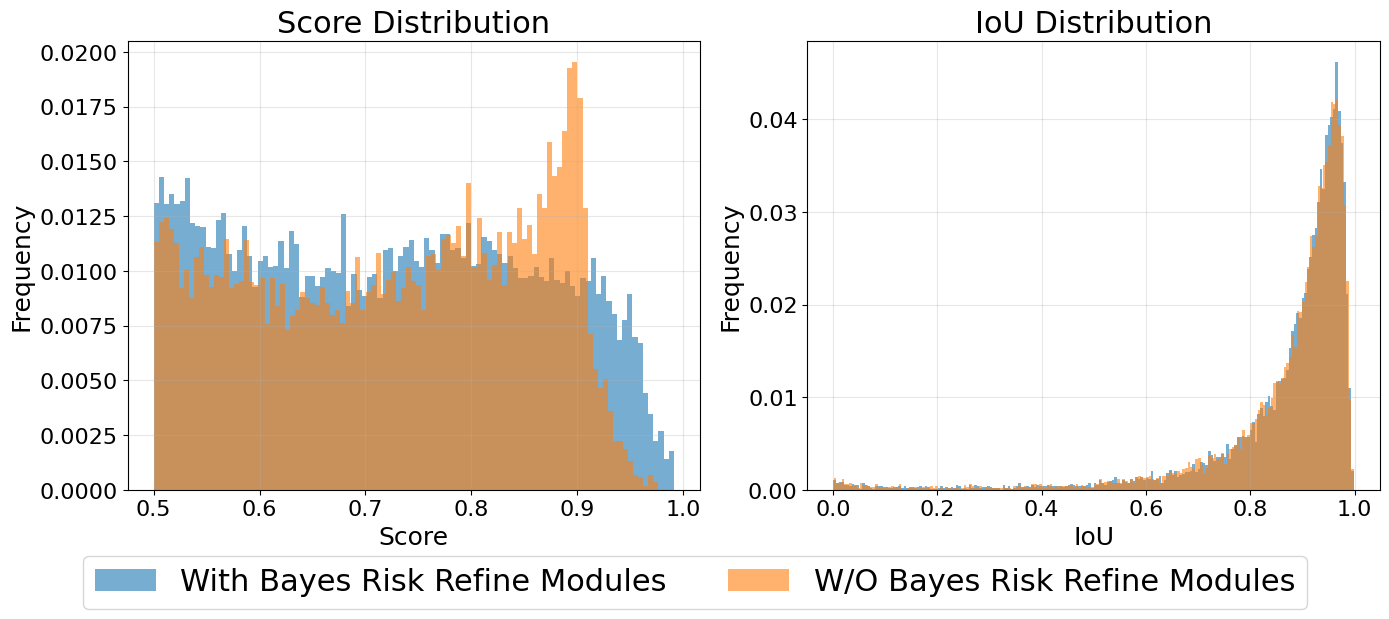}}
\vskip -0.1in
\caption{Density distribution of predicted classification scores and IoU before and after applying Bayes Risk Refinement Modules to the baseline model.}
\label{Role of BRRM}
\end{center}
\vskip -0.23in
\end{figure}

\textbf{Bayes Risk Refinement Modules. }The Bayes Risk Refinement Modules (BRRM) enhance model performance by adaptively refining predictions, prioritizing high-confidence regions while suppressing those with high Bayes Risk. As shown in \autoref{ablation_1}, BRRM improves overall performance, increasing AP by \textbf{0.8\%}, AP$_S$ by \textbf{0.7\%}, and AP$_M$ by \textbf{1.1\%}. \autoref{Role of BRRM} further illustrates its impact on score distribution. Before applying BRRM, the distribution is more peaked, indicating the model assigns excessive confidence to a concentrated set of predictions, making it more susceptible to incorrect high-confidence outputs. After applying BRRM, the distribution becomes smoother and more dispersed, reducing overconfidence and mitigating bias toward a few high-confidence scores by better incorporating Bayes Risk. Additionally, BRRM improves localization accuracy, as evidenced by a denser concentration in the high IoU range (\(>0.75\)) and a notable increase in extremely high IoU occurrences (\(\text{IoU} \approx 0.9\)).

\begin{table}[t]
\caption{The impact of incorporating each component across the Bayes Risk Refinement Modules. Here, BROE denotes Bayes Risk Output Embedding, BRMC represents Bayes Risk Matching Cost, and BRCL stands for Bayes Risk Classification Loss.}
\vspace{-1pt}
\label{ablation_3}
\begin{center}
\fontsize{9.5}{11}\selectfont  
\setlength{\tabcolsep}{2pt} 
\renewcommand{\arraystretch}{1} 
\begin{tabular}{ccccccccc} 
\toprule
BROE &  BRMC & BRCL & AP & AP$_{50}$ & AP$_{75}$ & AP$_S$ & AP$_M$ & AP$_L$ \\
\midrule
\xmark  & \xmark & \xmark & 49.3 & 66.9 & 53.6 & 32.4 & 52.7 & 63.7 \\
\cmark  & \xmark & \xmark & 49.5 & 67.0 & 54.1 & 32.6 & 52.9 & 63.6 \\
\cmark  & \cmark & \xmark & 49.9 & 67.3 & 54.6 & \textbf{33.2} & 53.4 & 63.9 \\
\cmark  & \cmark & \cmark & \textbf{50.1} & \textbf{67.6} & \textbf{54.8}& 33.1 & \textbf{53.8}& \textbf{64.0}\\
\bottomrule
\end{tabular}
\end{center}
\vspace{-3pt}
\end{table}

\begin{figure}[t]
\begin{center}
\centerline{\includegraphics[width=0.9\columnwidth]{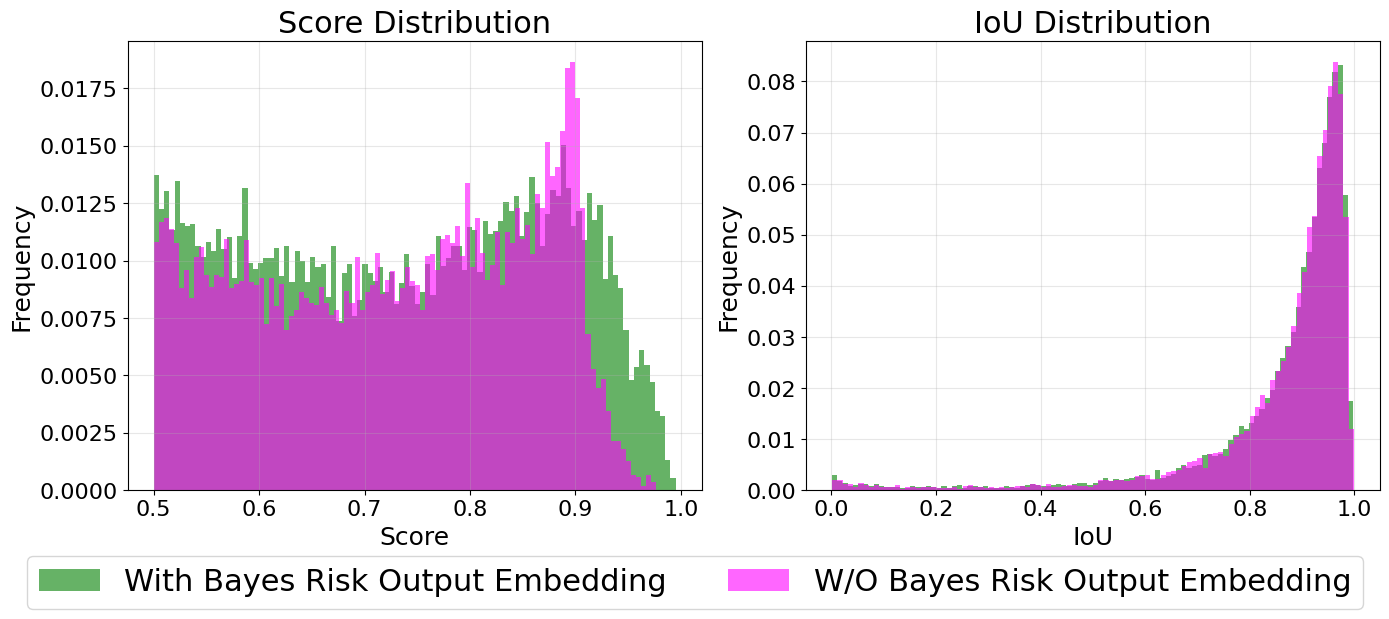}}
\vskip -0.1in
\caption{Density distribution of predicted classification scores and IoU before and after applying Bayes Risk Output Embedding to the baseline model.}
\label{Role of BROE}
\end{center}
\vskip -0.23in
\end{figure}

\textbf{Bayes Risk Output Embedding. }\autoref{ablation_3} and \autoref{ablation_2} demonstrate that incorporating Bayes Risk Output Embedding (BROE) improves performance. Specifically, adding BROE to the baseline model increases AP by \textbf{0.2\%} (\texttt{row1} vs. \texttt{row2} in Tab. \ref{ablation_3}), while adding BROE completes our approach, further boosting AP by \textbf{0.3\%} (\texttt{row1} vs. \texttt{row4} in Tab. \ref{ablation_2}). As shown in \autoref{Role of BROE}, after integrating BROE, the score distribution becomes smoother and more dispersed, and the IoU distribution shifts toward higher values, indicating improved localization accuracy and overall better prediction performance.

\begin{table}[!t]
\caption{The impact of removing each component in Bayes Risk Refinement Modules.}
\vspace{-1pt}
\label{ablation_2}
\begin{center}
\fontsize{9.5}{11}\selectfont  
\setlength{\tabcolsep}{2pt} 
\renewcommand{\arraystretch}{1} 
\begin{tabular}{ccccccccc} 
\toprule
BROE& BRMC &BRCL & AP & AP$_{50}$ & AP$_{75}$ & AP$_S$ & AP$_M$ & AP$_L$ \\
\midrule
\xmark &\cmark &\cmark  &49.8  &67.2  &54.5  &32.6  & 53.4 &63.8  \\
\cmark & \xmark &\cmark  & 49.7 & 67.2 & 54.0 & 32.0 & 53.0 & \textbf{64.2} \\
\cmark &\cmark &\xmark  & 49.9 & 67.3 & 54.6 & \textbf{33.2} & 53.4 & 63.9 \\
\cmark &\cmark &\cmark & \textbf{50.1} & \textbf{67.6} & \textbf{54.8} & 33.1 & \textbf{53.8} & 64.0 \\
\bottomrule
\end{tabular}
\end{center}
\vspace{-3pt}
\end{table}

\begin{figure}[!]
\begin{center}
\centerline{\includegraphics[width=0.75\columnwidth]{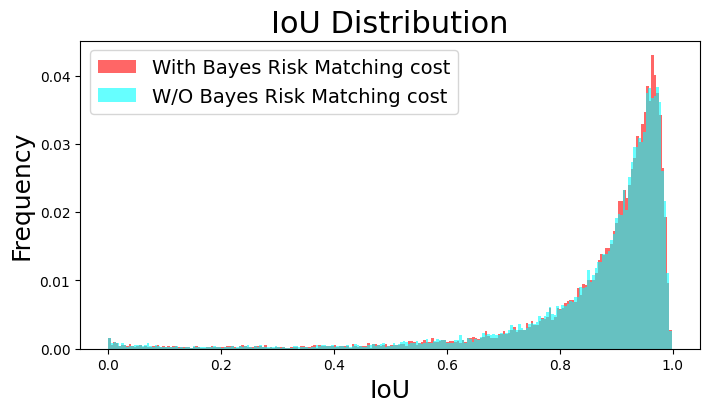}}
\vskip -0.1in
\caption{The density distribution of IoU before and after applying Bayes Risk Matching Cost to the baseline model with BROE.}
\label{Role  of BRMC}
\end{center}
\vskip -0.30in
\end{figure}

\textbf{Bayes Risk Matching Cost. }By integrating Bayes Risk with higher-order IoU formulations, the Bayes Risk Matching Cost (BRMC) better captures IoU variations, improving localization performance, particularly in AP$_S$. Specifically, BRMC increases AP by \textbf{0.4\%} and AP$_S$ by \textbf{0.6\%} (\texttt{row2} vs. \texttt{row3} in Tab. \ref{ablation_3}), and further boosts AP by \textbf{0.4\%} and AP$_S$ by \textbf{1.1\%} when completing our approach (\texttt{row2} vs. \texttt{row4} in Tab. \ref{ablation_2}). As shown in \autoref{Role  of BRMC}, BRMC enhances predicted bounding box alignment with ground truth, improving localization accuracy.

\textbf{Bayes Risk Classification Loss. }BRCL applies a Bayes Risk weighting mechanism to reduce the impact of high-Bayes-Risk regions, guiding the model to focus on more reliable areas. This encourages higher-confidence predictions with improved robustness. As shown in \autoref{ablation_2}, comparing \texttt{row3} and \texttt{row4}, incorporating BRCL further improves AP by \textbf{0.2\%}, even on an already strong-performing model.

\section{Conclusion}
In this paper, we address the limitations of conventional bounding box modeling in object detection by exploring a uncertainty-aware approach to enhance DETR-based methods. We model bounding boxes as Gaussian distributions to account for uncertainty and derive Bayes Risk to refine modules in DETRs. Moreover, we formulate the localization uncertainty for predictions. Extensive ablation studies and experimental results demonstrate that our method can be seamlessly integrated into existing DETRs, leading to improved performance. Our method can also be extended to domain-specific biomedical applications such as leukocyte detection.

\bibliographystyle{ACM-Reference-Format}
\bibliography{main}


\begin{thebibliography}{40}


\ifx \showCODEN    \undefined \def \showCODEN     #1{\unskip}     \fi
\ifx \showISBNx    \undefined \def \showISBNx     #1{\unskip}     \fi
\ifx \showISBNxiii \undefined \def \showISBNxiii  #1{\unskip}     \fi
\ifx \showISSN     \undefined \def \showISSN      #1{\unskip}     \fi
\ifx \showLCCN     \undefined \def \showLCCN      #1{\unskip}     \fi
\ifx \shownote     \undefined \def \shownote      #1{#1}          \fi
\ifx \showarticletitle \undefined \def \showarticletitle #1{#1}   \fi
\ifx \showURL      \undefined \def \showURL       {\relax}        \fi
\providecommand\bibfield[2]{#2}
\providecommand\bibinfo[2]{#2}
\providecommand\natexlab[1]{#1}
\providecommand\showeprint[2][]{arXiv:#2}

\bibitem[Cai et~al\mbox{.}(2023)]%
        {cai2023align-DETR}
\bibfield{author}{\bibinfo{person}{Zhi Cai}, \bibinfo{person}{Songtao Liu}, \bibinfo{person}{Guodong Wang}, \bibinfo{person}{Zheng Ge}, \bibinfo{person}{Xiangyu Zhang}, {and} \bibinfo{person}{Di Huang}.} \bibinfo{year}{2023}\natexlab{}.
\newblock \showarticletitle{Align-detr: Improving detr with simple iou-aware bce loss}.
\newblock \bibinfo{journal}{\emph{arXiv preprint arXiv:2304.07527}} (\bibinfo{year}{2023}).
\newblock


\bibitem[Carion et~al\mbox{.}(2020)]%
        {carion2020DETR}
\bibfield{author}{\bibinfo{person}{Nicolas Carion}, \bibinfo{person}{Francisco Massa}, \bibinfo{person}{Gabriel Synnaeve}, \bibinfo{person}{Nicolas Usunier}, \bibinfo{person}{Alexander Kirillov}, {and} \bibinfo{person}{Sergey Zagoruyko}.} \bibinfo{year}{2020}\natexlab{}.
\newblock \showarticletitle{End-to-end object detection with transformers}. In \bibinfo{booktitle}{\emph{European conference on computer vision}}. Springer, \bibinfo{pages}{213--229}.
\newblock


\bibitem[Chen et~al\mbox{.}(2024)]%
        {chen2024MFDS-DETR}
\bibfield{author}{\bibinfo{person}{Yifei Chen}, \bibinfo{person}{Chenyan Zhang}, \bibinfo{person}{Ben Chen}, \bibinfo{person}{Yiyu Huang}, \bibinfo{person}{Yifei Sun}, \bibinfo{person}{Changmiao Wang}, \bibinfo{person}{Xianjun Fu}, \bibinfo{person}{Yuxing Dai}, \bibinfo{person}{Feiwei Qin}, \bibinfo{person}{Yong Peng}, {et~al\mbox{.}}} \bibinfo{year}{2024}\natexlab{}.
\newblock \showarticletitle{Accurate leukocyte detection based on deformable-DETR and multi-level feature fusion for aiding diagnosis of blood diseases}.
\newblock \bibinfo{journal}{\emph{Computers in biology and medicine}}  \bibinfo{volume}{170} (\bibinfo{year}{2024}), \bibinfo{pages}{107917}.
\newblock


\bibitem[Choi et~al\mbox{.}(2019)]%
        {choi2019gaussian}
\bibfield{author}{\bibinfo{person}{Jiwoong Choi}, \bibinfo{person}{Dayoung Chun}, \bibinfo{person}{Hyun Kim}, {and} \bibinfo{person}{Hyuk-Jae Lee}.} \bibinfo{year}{2019}\natexlab{}.
\newblock \showarticletitle{Gaussian yolov3: An accurate and fast object detector using localization uncertainty for autonomous driving}. In \bibinfo{booktitle}{\emph{Proceedings of the IEEE/CVF International conference on computer vision}}. \bibinfo{pages}{502--511}.
\newblock


\bibitem[Dai et~al\mbox{.}(2021)]%
        {dai2021dynamic-DETR}
\bibfield{author}{\bibinfo{person}{Xiyang Dai}, \bibinfo{person}{Yinpeng Chen}, \bibinfo{person}{Jianwei Yang}, \bibinfo{person}{Pengchuan Zhang}, \bibinfo{person}{Lu Yuan}, {and} \bibinfo{person}{Lei Zhang}.} \bibinfo{year}{2021}\natexlab{}.
\newblock \showarticletitle{Dynamic detr: End-to-end object detection with dynamic attention}. In \bibinfo{booktitle}{\emph{Proceedings of the IEEE/CVF international conference on computer vision}}. \bibinfo{pages}{2988--2997}.
\newblock


\bibitem[Delon et~al\mbox{.}(2022)]%
        {delon2022gromovgwd-ours}
\bibfield{author}{\bibinfo{person}{Julie Delon}, \bibinfo{person}{Agnes Desolneux}, {and} \bibinfo{person}{Antoine Salmona}.} \bibinfo{year}{2022}\natexlab{}.
\newblock \showarticletitle{Gromov--Wasserstein distances between Gaussian distributions}.
\newblock \bibinfo{journal}{\emph{Journal of Applied Probability}} \bibinfo{volume}{59}, \bibinfo{number}{4} (\bibinfo{year}{2022}), \bibinfo{pages}{1178--1198}.
\newblock


\bibitem[Gevorgyan(2022)]%
        {gevorgyan2022SIoU}
\bibfield{author}{\bibinfo{person}{Zhora Gevorgyan}.} \bibinfo{year}{2022}\natexlab{}.
\newblock \showarticletitle{SIoU loss: More powerful learning for bounding box regression}.
\newblock \bibinfo{journal}{\emph{arXiv preprint arXiv:2205.12740}} (\bibinfo{year}{2022}).
\newblock


\bibitem[Guo and Zhang(2023)]%
        {guo2023YOLOv5-ALT}
\bibfield{author}{\bibinfo{person}{Yecai Guo} {and} \bibinfo{person}{Mengyao Zhang}.} \bibinfo{year}{2023}\natexlab{}.
\newblock \showarticletitle{Blood cell detection method based on improved YOLOv5}.
\newblock \bibinfo{journal}{\emph{IEEE Access}}  \bibinfo{volume}{11} (\bibinfo{year}{2023}), \bibinfo{pages}{67987--67995}.
\newblock


\bibitem[Harakeh et~al\mbox{.}(2020)]%
        {harakeh2020bayesod}
\bibfield{author}{\bibinfo{person}{Ali Harakeh}, \bibinfo{person}{Michael Smart}, {and} \bibinfo{person}{Steven~L Waslander}.} \bibinfo{year}{2020}\natexlab{}.
\newblock \showarticletitle{Bayesod: A bayesian approach for uncertainty estimation in deep object detectors}. In \bibinfo{booktitle}{\emph{2020 IEEE International Conference on Robotics and Automation (ICRA)}}. IEEE, \bibinfo{pages}{87--93}.
\newblock


\bibitem[He et~al\mbox{.}(2016)]%
        {he2016resnet}
\bibfield{author}{\bibinfo{person}{Kaiming He}, \bibinfo{person}{Xiangyu Zhang}, \bibinfo{person}{Shaoqing Ren}, {and} \bibinfo{person}{Jian Sun}.} \bibinfo{year}{2016}\natexlab{}.
\newblock \showarticletitle{Deep residual learning for image recognition}. In \bibinfo{booktitle}{\emph{Proceedings of the IEEE conference on computer vision and pattern recognition}}. \bibinfo{pages}{770--778}.
\newblock


\bibitem[He et~al\mbox{.}(2019a)]%
        {he2019bounding}
\bibfield{author}{\bibinfo{person}{Yihui He}, \bibinfo{person}{Chenchen Zhu}, \bibinfo{person}{Jianren Wang}, \bibinfo{person}{Marios Savvides}, {and} \bibinfo{person}{Xiangyu Zhang}.} \bibinfo{year}{2019}\natexlab{a}.
\newblock \showarticletitle{Bounding box regression with uncertainty for accurate object detection}. In \bibinfo{booktitle}{\emph{Proceedings of the ieee/cvf conference on computer vision and pattern recognition}}. \bibinfo{pages}{2888--2897}.
\newblock


\bibitem[He et~al\mbox{.}(2019b)]%
        {he2019bboxuncertainty}
\bibfield{author}{\bibinfo{person}{Yihui He}, \bibinfo{person}{Chenchen Zhu}, \bibinfo{person}{Jianren Wang}, \bibinfo{person}{Marios Savvides}, {and} \bibinfo{person}{Xiangyu Zhang}.} \bibinfo{year}{2019}\natexlab{b}.
\newblock \showarticletitle{Bounding box regression with uncertainty for accurate object detection}. In \bibinfo{booktitle}{\emph{Proceedings of the ieee/cvf conference on computer vision and pattern recognition}}. \bibinfo{pages}{2888--2897}.
\newblock


\bibitem[Hou et~al\mbox{.}(2025)]%
        {hou2025relation-DETR}
\bibfield{author}{\bibinfo{person}{Xiuquan Hou}, \bibinfo{person}{Meiqin Liu}, \bibinfo{person}{Senlin Zhang}, \bibinfo{person}{Ping Wei}, \bibinfo{person}{Badong Chen}, {and} \bibinfo{person}{Xuguang Lan}.} \bibinfo{year}{2025}\natexlab{}.
\newblock \showarticletitle{Relation detr: Exploring explicit position relation prior for object detection}. In \bibinfo{booktitle}{\emph{European Conference on Computer Vision}}. Springer, \bibinfo{pages}{89--105}.
\newblock


\bibitem[Jia et~al\mbox{.}(2023)]%
        {jia2023H-DETR}
\bibfield{author}{\bibinfo{person}{Ding Jia}, \bibinfo{person}{Yuhui Yuan}, \bibinfo{person}{Haodi He}, \bibinfo{person}{Xiaopei Wu}, \bibinfo{person}{Haojun Yu}, \bibinfo{person}{Weihong Lin}, \bibinfo{person}{Lei Sun}, \bibinfo{person}{Chao Zhang}, {and} \bibinfo{person}{Han Hu}.} \bibinfo{year}{2023}\natexlab{}.
\newblock \showarticletitle{Detrs with hybrid matching}. In \bibinfo{booktitle}{\emph{Proceedings of the IEEE/CVF conference on computer vision and pattern recognition}}. \bibinfo{pages}{19702--19712}.
\newblock


\bibitem[Lakshminarayanan et~al\mbox{.}(2017)]%
        {lakshminarayanan2017simple}
\bibfield{author}{\bibinfo{person}{Balaji Lakshminarayanan}, \bibinfo{person}{Alexander Pritzel}, {and} \bibinfo{person}{Charles Blundell}.} \bibinfo{year}{2017}\natexlab{}.
\newblock \showarticletitle{Simple and scalable predictive uncertainty estimation using deep ensembles}.
\newblock \bibinfo{journal}{\emph{Advances in neural information processing systems}}  \bibinfo{volume}{30} (\bibinfo{year}{2017}).
\newblock


\bibitem[Lee et~al\mbox{.}(2022)]%
        {lee2022localization}
\bibfield{author}{\bibinfo{person}{Youngwan Lee}, \bibinfo{person}{Joong-won Hwang}, \bibinfo{person}{Hyung-Il Kim}, \bibinfo{person}{Kimin Yun}, \bibinfo{person}{Yongjin Kwon}, \bibinfo{person}{Yuseok Bae}, {and} \bibinfo{person}{Sung~Ju Hwang}.} \bibinfo{year}{2022}\natexlab{}.
\newblock \showarticletitle{Localization uncertainty estimation for anchor-free object detection}. In \bibinfo{booktitle}{\emph{European Conference on Computer Vision}}. Springer, \bibinfo{pages}{27--42}.
\newblock


\bibitem[Li et~al\mbox{.}(2021)]%
        {li2021generalized}
\bibfield{author}{\bibinfo{person}{Xiang Li}, \bibinfo{person}{Wenhai Wang}, \bibinfo{person}{Xiaolin Hu}, \bibinfo{person}{Jun Li}, \bibinfo{person}{Jinhui Tang}, {and} \bibinfo{person}{Jian Yang}.} \bibinfo{year}{2021}\natexlab{}.
\newblock \showarticletitle{Generalized focal loss v2: Learning reliable localization quality estimation for dense object detection}. In \bibinfo{booktitle}{\emph{Proceedings of the IEEE/CVF conference on computer vision and pattern recognition}}. \bibinfo{pages}{11632--11641}.
\newblock


\bibitem[Lin et~al\mbox{.}(2017)]%
        {lin2017RetinaNet}
\bibfield{author}{\bibinfo{person}{Tsung-Yi Lin}, \bibinfo{person}{Priya Goyal}, \bibinfo{person}{Ross Girshick}, \bibinfo{person}{Kaiming He}, {and} \bibinfo{person}{Piotr Doll{\'a}r}.} \bibinfo{year}{2017}\natexlab{}.
\newblock \showarticletitle{Focal loss for dense object detection}. In \bibinfo{booktitle}{\emph{Proceedings of the IEEE international conference on computer vision}}. \bibinfo{pages}{2980--2988}.
\newblock


\bibitem[Lin et~al\mbox{.}(2014)]%
        {lin2014cocodata}
\bibfield{author}{\bibinfo{person}{Tsung-Yi Lin}, \bibinfo{person}{Michael Maire}, \bibinfo{person}{Serge Belongie}, \bibinfo{person}{James Hays}, \bibinfo{person}{Pietro Perona}, \bibinfo{person}{Deva Ramanan}, \bibinfo{person}{Piotr Doll{\'a}r}, {and} \bibinfo{person}{C~Lawrence Zitnick}.} \bibinfo{year}{2014}\natexlab{}.
\newblock \showarticletitle{Microsoft coco: Common objects in context}. In \bibinfo{booktitle}{\emph{Computer Vision--ECCV 2014: 13th European Conference, Zurich, Switzerland, September 6-12, 2014, Proceedings, Part V 13}}. Springer, \bibinfo{pages}{740--755}.
\newblock


\bibitem[Liu et~al\mbox{.}(2016)]%
        {liu2016ssd}
\bibfield{author}{\bibinfo{person}{Wei Liu}, \bibinfo{person}{Dragomir Anguelov}, \bibinfo{person}{Dumitru Erhan}, \bibinfo{person}{Christian Szegedy}, \bibinfo{person}{Scott Reed}, \bibinfo{person}{Cheng-Yang Fu}, {and} \bibinfo{person}{Alexander~C Berg}.} \bibinfo{year}{2016}\natexlab{}.
\newblock \showarticletitle{Ssd: Single shot multibox detector}. In \bibinfo{booktitle}{\emph{Computer Vision--ECCV 2016: 14th European Conference, Amsterdam, The Netherlands, October 11--14, 2016, Proceedings, Part I 14}}. Springer, \bibinfo{pages}{21--37}.
\newblock


\bibitem[Liu et~al\mbox{.}(2021)]%
        {liu2021swin-transformer}
\bibfield{author}{\bibinfo{person}{Ze Liu}, \bibinfo{person}{Yutong Lin}, \bibinfo{person}{Yue Cao}, \bibinfo{person}{Han Hu}, \bibinfo{person}{Yixuan Wei}, \bibinfo{person}{Zheng Zhang}, \bibinfo{person}{Stephen Lin}, {and} \bibinfo{person}{Baining Guo}.} \bibinfo{year}{2021}\natexlab{}.
\newblock \showarticletitle{Swin transformer: Hierarchical vision transformer using shifted windows}. In \bibinfo{booktitle}{\emph{Proceedings of the IEEE/CVF international conference on computer vision}}. \bibinfo{pages}{10012--10022}.
\newblock


\bibitem[Meng et~al\mbox{.}(2021)]%
        {meng2021conditional-DETR}
\bibfield{author}{\bibinfo{person}{Depu Meng}, \bibinfo{person}{Xiaokang Chen}, \bibinfo{person}{Zejia Fan}, \bibinfo{person}{Gang Zeng}, \bibinfo{person}{Houqiang Li}, \bibinfo{person}{Yuhui Yuan}, \bibinfo{person}{Lei Sun}, {and} \bibinfo{person}{Jingdong Wang}.} \bibinfo{year}{2021}\natexlab{}.
\newblock \showarticletitle{Conditional detr for fast training convergence}. In \bibinfo{booktitle}{\emph{Proceedings of the IEEE/CVF international conference on computer vision}}. \bibinfo{pages}{3651--3660}.
\newblock


\bibitem[Pu et~al\mbox{.}(2024)]%
        {pu2024rank-DETR}
\bibfield{author}{\bibinfo{person}{Yifan Pu}, \bibinfo{person}{Weicong Liang}, \bibinfo{person}{Yiduo Hao}, \bibinfo{person}{Yuhui Yuan}, \bibinfo{person}{Yukang Yang}, \bibinfo{person}{Chao Zhang}, \bibinfo{person}{Han Hu}, {and} \bibinfo{person}{Gao Huang}.} \bibinfo{year}{2024}\natexlab{}.
\newblock \showarticletitle{Rank-DETR for high quality object detection}.
\newblock \bibinfo{journal}{\emph{Advances in Neural Information Processing Systems}}  \bibinfo{volume}{36} (\bibinfo{year}{2024}).
\newblock


\bibitem[Qiu et~al\mbox{.}(2020)]%
        {qiu2020offset}
\bibfield{author}{\bibinfo{person}{Heqian Qiu}, \bibinfo{person}{Hongliang Li}, \bibinfo{person}{Qingbo Wu}, {and} \bibinfo{person}{Hengcan Shi}.} \bibinfo{year}{2020}\natexlab{}.
\newblock \showarticletitle{Offset bin classification network for accurate object detection}. In \bibinfo{booktitle}{\emph{Proceedings of the IEEE/CVF Conference on Computer Vision and Pattern Recognition}}. \bibinfo{pages}{13188--13197}.
\newblock


\bibitem[Redmon(2016)]%
        {redmon2016yolo1}
\bibfield{author}{\bibinfo{person}{J Redmon}.} \bibinfo{year}{2016}\natexlab{}.
\newblock \showarticletitle{You only look once: Unified, real-time object detection}. In \bibinfo{booktitle}{\emph{Proceedings of the IEEE conference on computer vision and pattern recognition}}.
\newblock


\bibitem[Ren et~al\mbox{.}(2016)]%
        {ren2016faster}
\bibfield{author}{\bibinfo{person}{Shaoqing Ren}, \bibinfo{person}{Kaiming He}, \bibinfo{person}{Ross Girshick}, {and} \bibinfo{person}{Jian Sun}.} \bibinfo{year}{2016}\natexlab{}.
\newblock \showarticletitle{Faster R-CNN: Towards real-time object detection with region proposal networks}.
\newblock \bibinfo{journal}{\emph{IEEE transactions on pattern analysis and machine intelligence}} \bibinfo{volume}{39}, \bibinfo{number}{6} (\bibinfo{year}{2016}), \bibinfo{pages}{1137--1149}.
\newblock


\bibitem[Rezatofighi et~al\mbox{.}(2019)]%
        {rezatofighi2019GIoU}
\bibfield{author}{\bibinfo{person}{Hamid Rezatofighi}, \bibinfo{person}{Nathan Tsoi}, \bibinfo{person}{JunYoung Gwak}, \bibinfo{person}{Amir Sadeghian}, \bibinfo{person}{Ian Reid}, {and} \bibinfo{person}{Silvio Savarese}.} \bibinfo{year}{2019}\natexlab{}.
\newblock \showarticletitle{Generalized intersection over union: A metric and a loss for bounding box regression}. In \bibinfo{booktitle}{\emph{Proceedings of the IEEE/CVF conference on computer vision and pattern recognition}}. \bibinfo{pages}{658--666}.
\newblock


\bibitem[Rezatofighi and Soltanian-Zadeh(2011)]%
        {rezatofighi2011LISC}
\bibfield{author}{\bibinfo{person}{Seyed~Hamid Rezatofighi} {and} \bibinfo{person}{Hamid Soltanian-Zadeh}.} \bibinfo{year}{2011}\natexlab{}.
\newblock \showarticletitle{Automatic recognition of five types of white blood cells in peripheral blood}.
\newblock \bibinfo{journal}{\emph{Computerized Medical Imaging and Graphics}} \bibinfo{volume}{35}, \bibinfo{number}{4} (\bibinfo{year}{2011}), \bibinfo{pages}{333--343}.
\newblock


\bibitem[Tian et~al\mbox{.}(2020)]%
        {tian2020fcos}
\bibfield{author}{\bibinfo{person}{Zhi Tian}, \bibinfo{person}{Chunhua Shen}, \bibinfo{person}{Hao Chen}, {and} \bibinfo{person}{Tong He}.} \bibinfo{year}{2020}\natexlab{}.
\newblock \showarticletitle{FCOS: A simple and strong anchor-free object detector}.
\newblock \bibinfo{journal}{\emph{IEEE transactions on pattern analysis and machine intelligence}} \bibinfo{volume}{44}, \bibinfo{number}{4} (\bibinfo{year}{2020}), \bibinfo{pages}{1922--1933}.
\newblock


\bibitem[Vaswani(2017)]%
        {vaswani2017transformer}
\bibfield{author}{\bibinfo{person}{A Vaswani}.} \bibinfo{year}{2017}\natexlab{}.
\newblock \showarticletitle{Attention is all you need}.
\newblock \bibinfo{journal}{\emph{Advances in Neural Information Processing Systems}} (\bibinfo{year}{2017}).
\newblock


\bibitem[Wang et~al\mbox{.}(2024a)]%
        {wang2024yolov10}
\bibfield{author}{\bibinfo{person}{Ao Wang}, \bibinfo{person}{Hui Chen}, \bibinfo{person}{Lihao Liu}, \bibinfo{person}{Kai Chen}, \bibinfo{person}{Zijia Lin}, \bibinfo{person}{Jungong Han}, {and} \bibinfo{person}{Guiguang Ding}.} \bibinfo{year}{2024}\natexlab{a}.
\newblock \showarticletitle{Yolov10: Real-time end-to-end object detection}.
\newblock \bibinfo{journal}{\emph{arXiv preprint arXiv:2405.14458}} (\bibinfo{year}{2024}).
\newblock


\bibitem[Wang et~al\mbox{.}(2024b)]%
        {wang2024gold}
\bibfield{author}{\bibinfo{person}{Chengcheng Wang}, \bibinfo{person}{Wei He}, \bibinfo{person}{Ying Nie}, \bibinfo{person}{Jianyuan Guo}, \bibinfo{person}{Chuanjian Liu}, \bibinfo{person}{Yunhe Wang}, {and} \bibinfo{person}{Kai Han}.} \bibinfo{year}{2024}\natexlab{b}.
\newblock \showarticletitle{Gold-YOLO: Efficient object detector via gather-and-distribute mechanism}.
\newblock \bibinfo{journal}{\emph{Advances in Neural Information Processing Systems}}  \bibinfo{volume}{36} (\bibinfo{year}{2024}).
\newblock


\bibitem[Wang et~al\mbox{.}(2023)]%
        {wang2023yolov7}
\bibfield{author}{\bibinfo{person}{Chien-Yao Wang}, \bibinfo{person}{Alexey Bochkovskiy}, {and} \bibinfo{person}{Hong-Yuan~Mark Liao}.} \bibinfo{year}{2023}\natexlab{}.
\newblock \showarticletitle{YOLOv7: Trainable bag-of-freebies sets new state-of-the-art for real-time object detectors}. In \bibinfo{booktitle}{\emph{Proceedings of the IEEE/CVF conference on computer vision and pattern recognition}}. \bibinfo{pages}{7464--7475}.
\newblock


\bibitem[Wang et~al\mbox{.}(2025)]%
        {wang2025yolov9}
\bibfield{author}{\bibinfo{person}{Chien-Yao Wang}, \bibinfo{person}{I-Hau Yeh}, {and} \bibinfo{person}{Hong-Yuan Mark~Liao}.} \bibinfo{year}{2025}\natexlab{}.
\newblock \showarticletitle{Yolov9: Learning what you want to learn using programmable gradient information}. In \bibinfo{booktitle}{\emph{European conference on computer vision}}. Springer, \bibinfo{pages}{1--21}.
\newblock


\bibitem[Wang et~al\mbox{.}(2021)]%
        {wang2021NWD}
\bibfield{author}{\bibinfo{person}{Jinwang Wang}, \bibinfo{person}{Chang Xu}, \bibinfo{person}{Wen Yang}, {and} \bibinfo{person}{Lei Yu}.} \bibinfo{year}{2021}\natexlab{}.
\newblock \showarticletitle{A normalized Gaussian Wasserstein distance for tiny object detection}.
\newblock \bibinfo{journal}{\emph{arXiv preprint arXiv:2110.13389}} (\bibinfo{year}{2021}).
\newblock


\bibitem[Xu et~al\mbox{.}(2022)]%
        {xu2022TE-YOLOF}
\bibfield{author}{\bibinfo{person}{Fanxin Xu}, \bibinfo{person}{Xiangkui Li}, \bibinfo{person}{Hang Yang}, \bibinfo{person}{Yali Wang}, {and} \bibinfo{person}{Wei Xiang}.} \bibinfo{year}{2022}\natexlab{}.
\newblock \showarticletitle{TE-YOLOF: Tiny and efficient YOLOF for blood cell detection}.
\newblock \bibinfo{journal}{\emph{Biomedical Signal Processing and Control}}  \bibinfo{volume}{73} (\bibinfo{year}{2022}), \bibinfo{pages}{103416}.
\newblock


\bibitem[Yang et~al\mbox{.}(2021)]%
        {yang2021GWD-old}
\bibfield{author}{\bibinfo{person}{Xue Yang}, \bibinfo{person}{Junchi Yan}, \bibinfo{person}{Qi Ming}, \bibinfo{person}{Wentao Wang}, \bibinfo{person}{Xiaopeng Zhang}, {and} \bibinfo{person}{Qi Tian}.} \bibinfo{year}{2021}\natexlab{}.
\newblock \showarticletitle{Rethinking rotated object detection with gaussian wasserstein distance loss}. In \bibinfo{booktitle}{\emph{International conference on machine learning}}. PMLR, \bibinfo{pages}{11830--11841}.
\newblock


\bibitem[Zhang et~al\mbox{.}(2022)]%
        {zhang2022dino-DETR}
\bibfield{author}{\bibinfo{person}{Hao Zhang}, \bibinfo{person}{Feng Li}, \bibinfo{person}{Shilong Liu}, \bibinfo{person}{Lei Zhang}, \bibinfo{person}{Hang Su}, \bibinfo{person}{Jun Zhu}, \bibinfo{person}{Lionel~M Ni}, {and} \bibinfo{person}{Heung-Yeung Shum}.} \bibinfo{year}{2022}\natexlab{}.
\newblock \showarticletitle{Dino: Detr with improved denoising anchor boxes for end-to-end object detection}.
\newblock \bibinfo{journal}{\emph{arXiv preprint arXiv:2203.03605}} (\bibinfo{year}{2022}).
\newblock


\bibitem[Zheng et~al\mbox{.}(2020)]%
        {zheng2020DCIoU}
\bibfield{author}{\bibinfo{person}{Zhaohui Zheng}, \bibinfo{person}{Ping Wang}, \bibinfo{person}{Wei Liu}, \bibinfo{person}{Jinze Li}, \bibinfo{person}{Rongguang Ye}, {and} \bibinfo{person}{Dongwei Ren}.} \bibinfo{year}{2020}\natexlab{}.
\newblock \showarticletitle{Distance-IoU loss: Faster and better learning for bounding box regression}. In \bibinfo{booktitle}{\emph{Proceedings of the AAAI conference on artificial intelligence}}, Vol.~\bibinfo{volume}{34}. \bibinfo{pages}{12993--13000}.
\newblock


\bibitem[Zhu et~al\mbox{.}(2020)]%
        {zhu2020Deformable-DETR}
\bibfield{author}{\bibinfo{person}{Xizhou Zhu}, \bibinfo{person}{Weijie Su}, \bibinfo{person}{Lewei Lu}, \bibinfo{person}{Bin Li}, \bibinfo{person}{Xiaogang Wang}, {and} \bibinfo{person}{Jifeng Dai}.} \bibinfo{year}{2020}\natexlab{}.
\newblock \showarticletitle{Deformable detr: Deformable transformers for end-to-end object detection}.
\newblock \bibinfo{journal}{\emph{arXiv preprint arXiv:2010.04159}} (\bibinfo{year}{2020}).
\newblock


\end{thebibliography}

\newpage
\appendix
\onecolumn
\section{Appendix}

\subsection{Ablation Experiments on GWD}Compared to traditional metrics such as IoU and GIoU, Gromov-Wasserstein distance and Wasserstein  distance measure the difference between the predicted and ground truth box distributions from a distributional perspective. We use the H-DETR model with IoU loss as the baseline and conduct experiments to demonstrate the impact of GIoU, Wasserstein distance and Gromov-Wasserstein distance.

As shown in \ref{table:ablation_gwd_IOU}, using Gromov-Wasserstein Distance and Wasserstein Distance provides greater performance improvement compared to GIoU. However, we will demonstrate in \ref{sec: WD} that the Wasserstein Distance formulation essentially relies on the same variables as GIoU and IoU, without incorporating covariance matrix terms that characterize the distribution of predicted boxes. Therefore, it cannot compute Bayes Risk for further model performance improvement.
\label{sec:appendix_gwd}
\begin{table*}[t]
\caption{Effect of GIoU, Wasserstein distance and Gromov-Wasserstein distance. Here $R = (c_x, c_y, w, h)$ represent the ground truth bbox, $\hat{R} = (\hat{c}_x, \hat{c}_y, \hat{w}, \hat{h})$ represent the predicted bbox.  }
\label{table:ablation_gwd_IOU}
\vskip 0.15in
\begin{center}
\begin{small} 
\begin{sc}
\resizebox{\textwidth}{!}{ 
\begin{tabular}{cccccccc}
\toprule
Metric Formulation  &  Type & AP & AP$_{50}$ & AP$_{75}$ & AP$_S$ & AP$_M$ & AP$_L$ \\
\midrule
$f(R,\hat{R})$ & IoU & 48.6 & 66.6 & 53.1 & 31.2 & 51.6 & 63.1 \\
$f(R,\hat{R})$ & GIoU & 48.7 & 66.4 & 52.9 & 31.2 & 51.5 & 63.5 \\
$f(R,\hat{R})$ & IoU+WD & 48.8 & 66.6 & 53.2 & 31.6 & 51.9 & 63.2 \\
$f(R,\hat{R},\sigma_{\hat{c}_x}^2,\sigma_{\hat{c}_y}^2,\sigma_{\hat{w}}^2,\sigma_{\hat{h}}^2)$ & IoU+GWD & 48.8 & 66.7 & 53.1 & 31.4 & 52.1 & 63.3 \\
\bottomrule
\end{tabular}
}
\end{sc}
\end{small} 
\end{center}
\vskip -0.1in
\end{table*}

\subsection{Formulation of Wasserstein distance}
\label{sec: WD}
As stated in \cite{wang2021NWD}, given the ground truth bounding box $R = (c_x, c_y, w, h)$ and the predicted bounding box $\hat{R} = (\hat{c}_x, \hat{c}_y, \hat{w}, \hat{h})$, the ground truth bbox follows a Gaussian distribution $m=\mathcal{N}_g(\boldsymbol{\mu}_g, \boldsymbol{\Sigma}_g) \in \mathbb{R}^2$, the predicted bbox follows another Gaussian distribution $n=\mathcal{N_{P}}(\boldsymbol{\mu}_\mathcal{P}, \boldsymbol{\Sigma}_\mathcal{P}) \in \mathbb{R}^2$, where 
\[
\boldsymbol{\mu}_g =
\begin{bmatrix}
c_x \\
c_y 
\end{bmatrix},
\quad
\boldsymbol{\Sigma}_g =
\begin{bmatrix}
w^2/4 & 0 \\
0 & h^2/4
\end{bmatrix},
\quad
\boldsymbol{\mu}_\mathcal{P} =
\begin{bmatrix}
\hat{c}_x\\
\hat{c}_y 
\end{bmatrix},
\quad
\boldsymbol{\Sigma}_\mathcal{P} =
\begin{bmatrix}
\hat{w}^2/4 & 0 \\
0 & \hat{h}^2/4
\end{bmatrix}.
\]
\begin{align}
    W_2^2(m, n) &= \|\boldsymbol{\mu}_g - \boldsymbol{\mu}_\mathcal{P}\|_2^2 + \operatorname{Tr} \left( \boldsymbol{\Sigma}_g + \boldsymbol{\Sigma}_\mathcal{P} - 2 \left( \boldsymbol{\Sigma}_\mathcal{P}^{1/2} \boldsymbol{\Sigma}_g \boldsymbol{\Sigma}_\mathcal{P}^{1/2} \right)^{1/2} \right) \notag \\
    &= \left\| 
    \begin{bmatrix} 
        c_x, c_y, \frac{w}{2}, \frac{h}{2} 
    \end{bmatrix}^{\mathrm{T}}
    - 
    \begin{bmatrix} 
        \hat{c}_x, \hat{c}_y, \frac{\hat{w}}{2}, \frac{\hat{h}}{2} 
    \end{bmatrix}^{\mathrm{T}}
    \right\|_2^2 \notag 
\end{align}

\subsection{Analytical solution for the Gromov-Wasserstein distance}
\label{Analytical solution}
Given $m=\mathcal{N}_g(\boldsymbol{\mu}_g, \boldsymbol{\Sigma}_g)$ and $n=\mathcal{N_{P}}(\boldsymbol{\mu}_\mathcal{P}, \boldsymbol{\Sigma}_\mathcal{P})$, as stated in \cite{delon2022gromovgwd-ours}, Gromov-Wasserstein distance between the ground truth bbox distribution and predicted 
bbox distribution has the following expression:

$GGW_2^2(m, n) = 4 \left( \text{tr}(\boldsymbol{\Sigma}_\mathcal{P}) - \text{tr}(\boldsymbol{\Sigma}_g) \right)^2 + 8 \left\| \boldsymbol{\Sigma}_\mathcal{P}^{(2)} - \boldsymbol{\Sigma}_g \right\|_F^2 + 8 \left( \left\|\boldsymbol{\Sigma}_\mathcal{P}\right\|_F^2 - \left\|\boldsymbol{\Sigma}_\mathcal{P}^{(2)} \right\|_F^2 \right)$, where $\boldsymbol{\Sigma}_\mathcal{P}^{(2)}$ denotes the submatrix containing the $2$ first row and the $2$ first columns of $\boldsymbol{\Sigma}_\mathcal{P}$.
 
\subsection{Proof of \autoref{thm:convergence}}
\label{proof of convergence}
We first need the following lemma to illustrate the situation when $GGW_2^2(m, n)=0$.
\begin{lemma}
\label{lem:GW=0}
Given $m=\mathcal{N}_g(\boldsymbol{\mu}_g, \boldsymbol{\Sigma}_g)$ and $n=\mathcal{N_{P}}(\boldsymbol{\mu}_\mathcal{P}, \boldsymbol{\Sigma}_\mathcal{P})$, where $m \in \mathbb{R}^2$ and $n \in \mathbb{R}^4$ , in this case, 
$GGW_2^2(m, n)=0$ when $
\boldsymbol{\Sigma}_\mathcal{P}=
\begin{pmatrix}
\boldsymbol{\Sigma}_g & \boldsymbol{0} \\
\boldsymbol{0} & \boldsymbol{0}
\end{pmatrix} \in \mathbb{R}^4
$.

\end{lemma}

We are ready to prove \ref{thm:convergence} now.

\begin{proof} 
Since we have the analytical solution for the Gromov-Wasserstein distance as: 
\begin{align}
GGW_2^2(m, n) = &\ 4 \left( \text{tr}(\boldsymbol{\Sigma}_\mathcal{P}) - \text{tr}(\boldsymbol{\Sigma}_g) \right)^2 \notag 
+ 8 \left\| \boldsymbol{\Sigma}_\mathcal{P}^{(2)} - \boldsymbol{\Sigma}_g \right\|_F^2 \notag 
+ 8 \left( \left\|\boldsymbol{\Sigma}_\mathcal{P}\right\|_F^2 - \left\|\boldsymbol{\Sigma}_\mathcal{P}^{(2)} \right\|_F^2 \right).
\end{align} and it can been seem through \ref{lem:GW=0} that $GGW_2^2(m, n)=0$, when $
\boldsymbol{\Sigma}_\mathcal{P}=\boldsymbol{\Sigma}_*=
\begin{pmatrix}
\boldsymbol{\Sigma}_g & \boldsymbol{0} \\
\boldsymbol{0} & \boldsymbol{0}
\end{pmatrix} \in \mathbb{R}^4
$. 

For the normal case, when $GGW_2^2(m, n)$ is approaching zero,
we  consider \( \boldsymbol{\Sigma}_\mathcal{P} \) has the form that \( \boldsymbol{\Sigma}_\mathcal{P} = \boldsymbol{\Sigma}_* +\Delta \boldsymbol{\Sigma} \),  where \( \Delta \boldsymbol{\Sigma} \) is a small perturbation matrix such that $\|\Delta \boldsymbol{\Sigma}\|_F \to 0$.

Now we consider the first term in the expression:  $ \left( \text{tr}(\boldsymbol{\Sigma}_\mathcal{P}) - \text{tr}(\boldsymbol{\Sigma}_g) \right)^2 $ 
\begin{align}
 \left( \text{tr}(\boldsymbol{\Sigma}_\mathcal{P}) - \text{tr}(\boldsymbol{\Sigma}_g) \right)^2 
 = &\ (\text{tr}(\Delta \boldsymbol{\Sigma}))^2 \notag
\end{align} Notice that $\left| \text{tr}(\Delta \boldsymbol{\Sigma}) \right| \leq \|\Delta \boldsymbol{\Sigma}\|_F
$, thus we have  \begin{align} \left( \text{tr}(\boldsymbol{\Sigma}_\mathcal{P}) - \text{tr}(\boldsymbol{\Sigma}_g) \right)^2 =
(\text{tr}(\Delta \boldsymbol{\Sigma}))^2 
 =  &\ O( {\|\Delta \boldsymbol{\Sigma}\|}^2_F) \notag
\end{align}
For the second term in the expression: $ \left\| \boldsymbol{\Sigma}_\mathcal{P}^{(2)} - \boldsymbol{\Sigma}_g \right\|_F^2$

 \begin{align} \left\| \boldsymbol{\Sigma}_\mathcal{P}^{(2)} - \boldsymbol{\Sigma}_g \right\|_F^2 = &\ {\|\Delta \boldsymbol{\Sigma}^{(2)}\|}^2_F = O( {\|\Delta \boldsymbol{\Sigma}\|}^2_F) \notag
\end{align}
For the last term in the expression: $  \left\|\boldsymbol{\Sigma}_\mathcal{P}\right\|_F^2 - \left\|\boldsymbol{\Sigma}_\mathcal{P}^{(2)} \right\|_F^2  $,

\begin{align}
\|\boldsymbol{\Sigma}_\mathcal{P}\|_F^2 &= \|\boldsymbol{\Sigma}_* + \Delta \boldsymbol{\Sigma}\|_F^2  
= \|\boldsymbol{\Sigma}_*\|_F^2 + 2\text{tr}(\boldsymbol{\Sigma}_*^\top \Delta \boldsymbol{\Sigma}) + \|\Delta \boldsymbol{\Sigma}\|_F^2 \notag
\end{align}

\begin{align}
\|\boldsymbol{\Sigma}_\mathcal{P}^{(2)}\|_F^2 &= \|\boldsymbol{\Sigma}_g + \Delta \boldsymbol{\Sigma}^{(2)}\|_F^2  
= \|\boldsymbol{\Sigma}_g \|_F^2 + 2\text{tr}(\boldsymbol{\Sigma}_g ^\top \Delta \boldsymbol{\Sigma}) + \|\Delta \boldsymbol{\Sigma}^{(2)}\|_F^2 \notag
\end{align}
Notice that $\text{tr}(\boldsymbol{\Sigma}_*^\top \Delta \boldsymbol{\Sigma}) = \text{tr}(\boldsymbol{\Sigma}_g ^\top \Delta \boldsymbol{\Sigma})$, so we have
\begin{align}
\|\boldsymbol{\Sigma}_\mathcal{P}\|_F^2 - \|\boldsymbol{\Sigma}_\mathcal{P}^{(2)}\|_F^2 
&= \|\Delta \boldsymbol{\Sigma}\|_F^2 - \|\Delta \boldsymbol{\Sigma}^{(2)}\|_F^2 = O( {\|\Delta \boldsymbol{\Sigma}\|}^2_F) \notag
\end{align}
Combining above three terms together, we have 
\begin{align}
\left( \text{tr}(\boldsymbol{\Sigma}_\mathcal{P}) - \text{tr}(\boldsymbol{\Sigma}_g) \right)^2 +
\left\| \boldsymbol{\Sigma}_\mathcal{P}^{(2)} - \boldsymbol{\Sigma}_g \right\|_F^2 +
\|\boldsymbol{\Sigma}_\mathcal{P}\|_F^2 - \|\boldsymbol{\Sigma}_\mathcal{P}^{(2)}\|_F^2 
& = O( {\|\Delta \boldsymbol{\Sigma}\|}^2_F) \notag
\end{align}
so we have $GW_2^2(m, n) = O( {\|\Delta \boldsymbol{\Sigma}\|}^2_F)$.

\end{proof}

\subsection{Derivation of Bayes Risk for $L_2$ loss}

\label{Bayes Risk derivation}
Let \( R = (c_x, c_y, w, h) \) denotes the ground truth bounding box and  \( \hat{R} = (\hat{c}_x, \hat{c}_y, \hat{w}, \hat{h}) \) denotes the predicted bounding box. For bounding box regression task, we choose $L_2$ loss as the loss function to reflect the difference between those two bounding boxes. The $L_2$ loss  has the following formulation :
\begin{align}
L_2 = (\hat{c}_x - c_x)^2 + (\hat{c}_y - c_y)^2 + (\hat{w} - w)^2 + (\hat{h} - h)^2 \notag
\end{align}
Notice that for the predicted bounding box \( \hat{R} = (\hat{c}_x, \hat{c}_y, \hat{w}, \hat{h}) \), it  follows a 4D Gaussian distribution $\mathcal{N_{P}}(\boldsymbol{\mu}_\mathcal{P}, \boldsymbol{\Sigma}_\mathcal{P})$ with 
\[
\boldsymbol{\mu}_\mathcal{P} =
\begin{bmatrix}
c_x \\
c_y \\
w \\
h
\end{bmatrix},
\quad
\boldsymbol{\Sigma}_\mathcal{P} =
\begin{bmatrix}
\sigma_{\hat{c}_x}^2 & 0 & 0 & 0 \\
0 & \sigma_{\hat{c}_y}^2 & 0 &  0 \\
0 & 0 & \sigma_{\hat{w}}^2 & 0 \\
0 & 0 & 0 & \sigma_{\hat{h}}^2
\end{bmatrix}.
\]
Each component of the ground truth bounding box  \( R = (c_x, c_y, w, h) \) follows a uniform distribution $U(0,1)$ on the interval [0,1].

Take $\hat{c}_x$ for illustration, it then has the posterior distribution $p(\hat{c}_x|c_x) = \frac{1}{\sqrt{2\pi \sigma_{\hat{c}_x}^2}} \exp\left(-\frac{(x - c_x)^2}{2\sigma_{\hat{c}_x}^2}\right) $ and $c_x \sim U(0,1)$, so the Bayes Risk can be calculated as:
\begin{align}
\text{Bayes Risk for } \hat{c}_x &= \iint (\hat{c}_x - c_x)^2 p(\hat{c}_x, c_x) \, d\hat{c}_x \, dc_x \notag \\ 
&= \iint (\hat{c}_x - c_x)^2 p(\hat{c}_x | c_x) p(c_x) \, d\hat{c}_x \, dc_x \notag \\
&= \int \left[ \int (\hat{c}_x - c_x)^2 p(\hat{c}_x | c_x)\, d\hat{c}_x \right] p(c_x) \, dc_x \notag \\
&= \int \sigma_{\hat{c}_x}^2 p(c_x) \, dc_x \notag \\
&= \sigma_{\hat{c}_x}^2 \notag
\end{align}
So the Bayes Risk for \( \hat{R} = (\hat{c}_x, \hat{c}_y, \hat{w}, \hat{h}) \) is:
\begin{align}
\textit{Risk}^* = \sigma_{\hat{c}_x}^2 + \sigma_{\hat{c}_y}^2 + \sigma_{\hat{w}}^2  + \sigma_{\hat{h}}^2 \notag
\end{align}

\subsection{Adapting Our Method to YOLO}
To show that our modeling approach generalizes beyond DETR, we applied it to the YOLOv5 model, the results are shown in \autoref{tab:yolo}. Due to YOLO’s one-stage architecture, the Bayes Risk refinement module (BRRM) here only modifies the output embeddings and the classification loss.
\begin{table*}[t]
    \centering
    \caption{Results on a smaller-scale VOC dataset.}
    \fontsize{10.5}{12}\selectfont  
    \setlength{\tabcolsep}{4pt} 
    \renewcommand{\arraystretch}{1.1} 
    \begin{tabular}{c|c|c|c|c}
    \toprule
    Method & AP & AP$_S$ & AP$_M$ & AP$_L$ \\
    \midrule
    YOLOV5 & 51.5 & 24.6 & 39.9 & 56.9 \\
    YOLOV5+WD & 51.7 & 24.8 & 40.7 & 56.8 \\
    YOLOV5+GWD & 51.9 & 25.0& 41.0 & 57.2 \\
    \textbf{YOLOV5+GWD+BRRM (Ours)} & \textbf{52.4} & \textbf{25.3} & \textbf{41.6} & \textbf{57.8}\\
    \bottomrule
    \end{tabular}
    \label{tab:yolo}
\end{table*}

\subsection{Analysis of Parameter k in Algorithm 1}
\autoref{tab:para k} shows the effect of division count k on localization‐uncertainty computation. To highlight these effects, we report total inference time and the standard deviation of the estimated uncertainty on the COCO test set.
Inference time grows linearly with k, while uncertainty’s standard deviation decreases—indicating more robust estimates. When k exceeds 300, robustness gains plateau, so we choose k = 300 as a trade‐off between efficiency and estimation quality.

\begin{table}
    \centering
    \caption{Impact of Parameter k on COCO.}
    \fontsize{10.5}{12}\selectfont  
    \setlength{\tabcolsep}{4pt} 
    \renewcommand{\arraystretch}{1.1} 
    \begin{tabular}{l|cccccc}
    \toprule
     K & 100 & 200 & \textbf{300} & 400 & 500 & 600 \\
    \midrule
     Inference time(min)   & 5.13 & 6.95 & \textbf{8.68} & 10.97 & 13.03 & 14.67 \\
     Std of uncertainty   & 0.106 & 0.085 & \textbf{0.069} & 0.064 & 0.060 & 0.058 \\
    \bottomrule
    \end{tabular}
    \label{tab:para k}
\end{table}

\subsection{Computational Complexity Analysis}
\begin{itemize}
    \item \textbf{Theoretical Analysis:} The extra computational overhead comes from: (1) GWD computation—O(N). (2) Quantifying localization uncertainty (Alg. 1)—O(N·k), where N is the number of boxes, k is the number of divisions.
    \item \textbf{Experimental Analysis:} We train H-DETR and our enhanced model on COCO using eight 3090 GPUs(12 epochs) and recorded training time. We also measure single-image inference time. \autoref{tab:time} shows that added computation introduced by our method remains practically acceptable.

    \begin{table}
    \centering
    \caption{Time Consumption on COCO.}
    \fontsize{10.5}{12}\selectfont  
    \setlength{\tabcolsep}{4pt} 
    \renewcommand{\arraystretch}{1.1} 
    \begin{tabular}{l|ccc}
    \toprule
       & H-DETR & H-DETR+Ours & Rise \\
    \midrule
     Training time(h)   & 14.67 & 15.55 & 6\% \\
     Inference time(h)  & 0.24 & 0.25 & 4.16\% \\
    \bottomrule
    \end{tabular}
    \label{tab:time}
\end{table}
\end{itemize}

\end{document}